\documentclass[journal,10pt]{IEEEtai}
\ifCLASSINFOpdf
\usepackage[pdftex]{graphicx}
\graphicspath{{../pdf/}{../jpeg/}}
\DeclareGraphicsExtensions{.pdf,.jpeg,.png}
\else
\fi
\usepackage{etoolbox}
\usepackage
{float}
\usepackage[english]{babel}
\usepackage{adjustbox}
\usepackage{amssymb}
\usepackage{amsmath}
\usepackage{amsthm}
\usepackage{mathtools}

\usepackage{relsize}
\usepackage{lineno}
\usepackage{ellipsis}
\usepackage{textcomp}
\usepackage{cite}
\usepackage{graphicx}  
\usepackage{subfigure}

\usepackage{epsfig}
\usepackage{epstopdf}
\usepackage{caption}
\usepackage{psfrag}
\usepackage{grffile}

\usepackage{color}
\usepackage{hyperref}
\usepackage[english]{babel}
\usepackage{amsthm}
\usepackage{listings}
\usepackage[ruled,vlined]{algorithm2e}

\SetCommentSty{mycommfont}

\usepackage{xcolor}
\definecolor{codegreen}{rgb}{0,0.6,0}
\definecolor{codegray}{rgb}{0.5,0.5,0.5}
\definecolor{codepurple}{rgb}{0.58,0,0.82}
\definecolor{backcolour}{rgb}{0.95,0.95,0.92}
\lstdefinestyle{mystyle}{
  backgroundcolor=\color{backcolour},   commentstyle=\color{codegreen},
  keywordstyle=\color{magenta},
  numberstyle=\tiny\color{codegray},
  stringstyle=\color{codepurple},
  basicstyle=\ttfamily\footnotesize,
  breakatwhitespace=false,         
  breaklines=true,                 
  captionpos=b,                    
  keepspaces=true,                 
  numbers=left,                    
  numbersep=5pt,                  
  showspaces=false,                
  showstringspaces=false,
  showtabs=false,                  
  tabsize=2
}
\usepackage{mathtools}
\usepackage{relsize}
\usepackage{lineno}
\usepackage{makecell}

\usepackage{array}
\usepackage{parskip}
\usepackage{array}
\newcolumntype{L}[1]{>{\raggedright\let\newline\\\arraybackslash\hspace{0pt}}m{#1}}
\newcolumntype{C}[1]{>{\centering\let\newline\\\arraybackslash\hspace{0pt}}m{#1}}
\newcolumntype{R}[1]{>{\raggedleft\let\newline\\\arraybackslash\hspace{0pt}}m{#1}}

\usepackage{multirow}
\usepackage[most]{tcolorbox}
\tcbset{
  coltitle=black, fonttitle=\bfseries, sharp corners, enhanced, breakable,
  boxsep=1mm, left=1.8mm, right=1.8mm, top=1.2mm, bottom=1.2mm
}
\allowdisplaybreaks

\newcommand{\Dist}{\delta}             

\usepackage[T1]{fontenc}

\usepackage[utf8]{inputenc}

\usepackage{microtype}

\usepackage{inconsolata}

\vspace{\baselineskip}
\AtEndEnvironment{theorem}{\null\hfill\qedsymbol}

\newtcolorbox{promptbox}[1][]{colback=gray!3,colframe=gray!20,#1}
\begin{document}

\title{\textit{Faithful or Fabricated?} A Causal Framework for Rationalization Bias in LLM Judges}


	

\author{
\begin{tabular}{c}
Riya Tapwal\\
School of Computing and Electrical Engineering\\
Indian Institute of Technology (IIT) Mandi\\
\texttt{riya@iitmandi.ac.in}
\end{tabular}
\and
\begin{tabular}{c}
Abhishek Kumar\\
The Alan Turing Institute, London, U.K.\\
\texttt{akumar@turing.ac.uk}
\end{tabular}
\and
\begin{tabular}{c}
Carsten Maple\\
Warwick Manufacturing Group\\
University of Warwick, U.K.\\
\texttt{CM@warwick.ac.uk}
\end{tabular}
}

\maketitle

\begin{abstract}
Large language models (LLMs) are increasingly used as automatic judges for summarization and dialogue evaluation. Prior work has documented biases such as position, verbosity, and style preferences, but largely focuses on \emph{outcomes}, leaving judge \emph{explanations} underexplored. We instead ask whether LLM judges are \textbf{cue-invariant}, i.e., whether their rankings and explanations remain stable when \emph{non-evidential cues} are perturbed while holding the underlying texts fixed. We introduce a suite of \emph{cue interventions} (Blind, Truth, Flip, Placebo, Reveal-After) and tie-aware metrics that quantify \emph{outcome anchoring} and \emph{rationale anchoring} (label-aligned rhetoric and explanation drift), alongside consistency and stereotype-intrusion checks. We design \emph{anchoring attacks} via verbosity and confidence cues, and compare two mitigations: structured chain-of-thought prompting and \textsc{Proof-Before-Preference} (evidence lock $\rightarrow$ score $\rightarrow$ rank). Using a new dataset of 1{,}000 summaries from traditional extractive models and LLMs, we find substantial \emph{cue-anchored rationalization} under label/placebo perturbations, while \textsc{Proof-Before-Preference} markedly improves cue invariance over baselines.
\end{abstract}

\begin{IEEEImpStatement}
The increasing adoption of large language models (LLMs) as automated judges in evaluation pipelines raises critical concerns about the reliability and faithfulness of their decisions and explanations. This work makes a timely and impactful contribution by introducing a causal framework to formally characterize and quantify rationalization bias, where LLM judges align their verdicts and explanations with non-evidential cues rather than the underlying textual evidence. By proposing cue-invariance probing, anchoring metrics, and the Proof-Before-Preference (PBP) mitigation protocol, this study provides both diagnostic tools and practical solutions to improve robustness, fairness, and auditability in LLM-based evaluation systems. These advances are particularly significant for high-stakes applications such as benchmarking, compliance monitoring, and automated decision support, where unreliable or cue-sensitive judgments could undermine trust and fairness. At the same time, this work promotes responsible deployment by exposing systemic vulnerabilities and demonstrating mitigation strategies that reduce post-hoc rationalization, thereby contributing to the development of more transparent, accountable, and trustworthy AI systems.
\end{IEEEImpStatement}


\section{Introduction}

Summarization has long been a flagship task in NLP, historically benchmarked against human-authored gold summaries and judged by human annotators \cite{INR-015}. Yet this paradigm is rapidly disappearing. Do humans still summarize? In practice, the answer is largely no \cite{10.1162/tacl_a_00373}. Producing human summaries at scale is prohibitively costly and inconsistent, while automated systems, both traditional extractive methods and modern large language models (LLMs), can generate summaries instantly \cite{10.5555/3524938.3525989, 10.1162/tacl_a_00373, INR-015}. Further, relying on human evaluation does not scale, leading to the widespread adoption of LLMs as judges \cite{Desmond2025EvalAssist}. In this new regime, the central question shifts from whether LLMs match human summarization quality to whether our \emph{evaluation pipeline} remains reliable when decisions and explanations are delegated to LLMs. In particular, applications that still favor lightweight extractive summarization, e.g., large-scale monitoring, enterprise reporting, or compliance, depend on judges whose verdicts and rationales are grounded in the \emph{evidence}, not in superficial artifacts. Prior studies have documented outcome-level biases in LLM judges, including position, verbosity, stylistic preference, and self-enhancement \cite{Turpin2023UnfaithfulCoT}. These findings are important but incomplete: they primarily track \emph{what} the judge decides, not \emph{why}. A decision is trustworthy only if its explanation reflects the same evidential features of the input that drove the choice. If, instead, explanations realign to extraneous signals, labels, badges, or stylistic hints, then evaluation can be steered without changing the underlying texts, eroding both auditability and fairness.

\begin{figure*}
    \centering
    \includegraphics[width=0.65\textwidth]{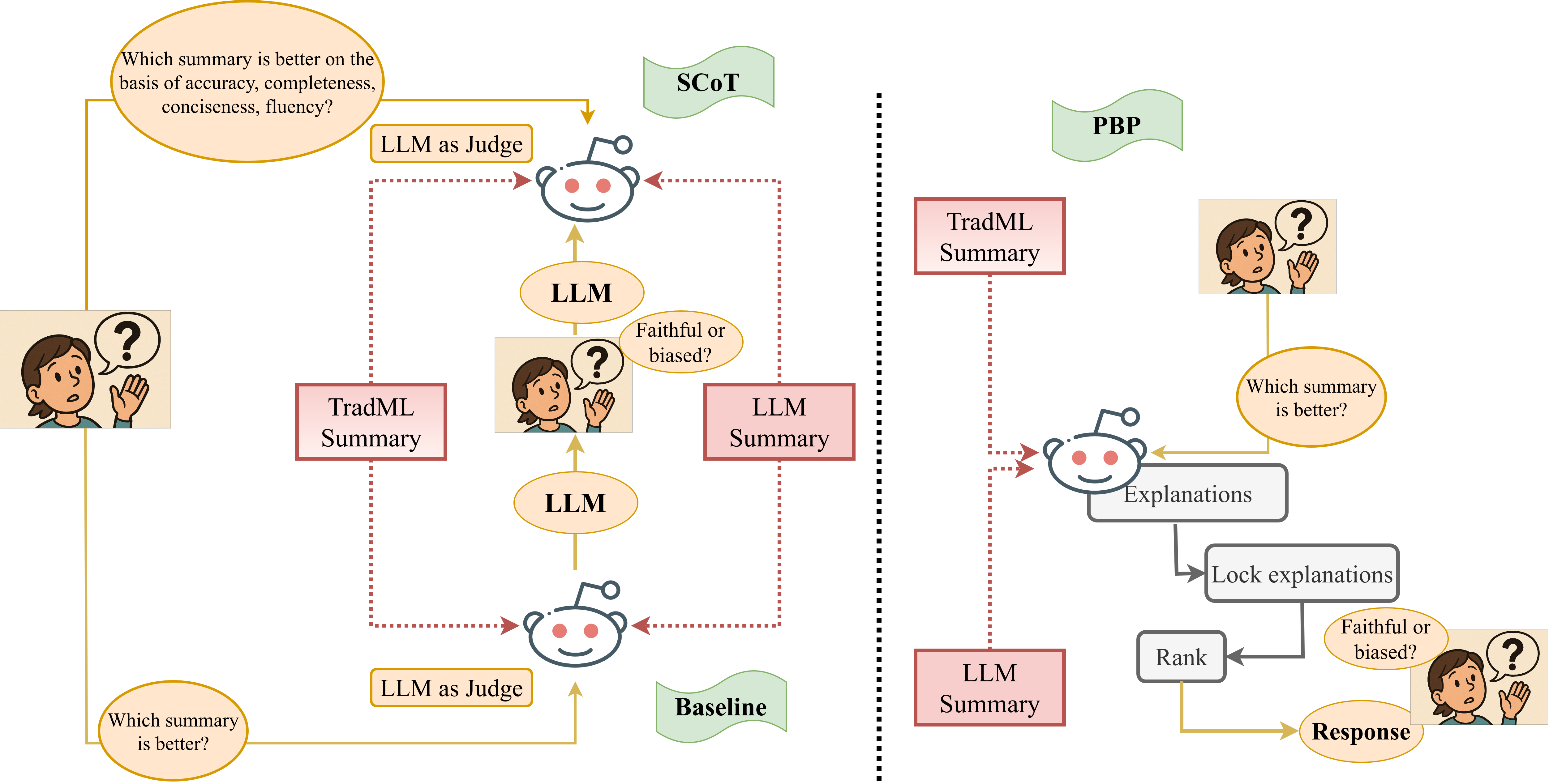}
    \caption{Overview of the three judging protocols and where rationalization can arise.
\emph{Baseline} (left: bottom): a single LLM judge directly chooses between an LLM and a TradML summary; explanations are post-hoc and thus cue/label-susceptible.
\emph{SCoT} (left:top): the judge reasons along predefined criteria (accuracy, completeness, conciseness, fluency) before deciding, but evidence is not locked, allowing rubric-amplified rationalization.
\emph{PBP} (right): Proof-Before-Preference, the judge first writes and \emph{locks} criterion-wise evidence, then scores and aggregates to rank, which curbs label anchoring and explanation drift.}

    \label{Fig 8}
\end{figure*}

We frame this reliability requirement as \textbf{cue invariance}. Let $X$ denote the fixed textual evidence (source document and candidate summaries) and let $C$ denote \emph{non-evidential cues} such as metadata labels. An LLM judge is cue-invariant if its ranking $r$ and explanation $e$ remain stable when $C$ is perturbed while $X$ is held constant. This perspective turns a vague notion of ``faithfulness'' into a precise robustness target: measure the causal effect of $C$ on $(r,e)$ under controlled interventions. To that end, we introduce a suite of \emph{cue interventions} that manipulate $C$ while keeping $X$ fixed: \textsc{Blind} (no cues), \textsc{Truth} (ground-truth label), \textsc{Flip} (inverted label), \textsc{Placebo} (credible but non-informative badge), and \textsc{Reveal-After} (applies FLIP, then reveals TRUTH). These probes expose when a judge’s decisions and explanations move toward the presented cues. We complement them with \emph{tie-aware} metrics that separate two phenomena: \emph{outcome anchoring} (directional shifts in rankings) and \emph{rationale anchoring} (label-aligned rhetoric and explanation drift). The same framework also reveals a structural weakness: \emph{anchoring attacks} that use innocuous style cues, verbosity and confidence, to sway both decisions and rationales even for identical summaries. Finally, we study defenses. A \emph{criterion-guided structured chain-of-thought} (SCoT) prompts judges to reason along explicit dimensions (accuracy, completeness, conciseness, fluency) before deciding. Building on this, \emph{\textsc{Proof-Before-Preference} (PBP)} first \emph{locks} criterion-wise notes with cited spans, then scores and ranks strictly from the locked evidence, reducing opportunities for post-hoc rationalization and label anchoring.

\paragraph{Motivation}

The increasing use of large language models (LLMs) as automated judges in evaluation pipelines has introduced significant scalability and efficiency benefits, but it also raises critical concerns about the reliability and faithfulness of their decisions and explanations. While prior work has identified outcome-level biases such as position, verbosity, and stylistic preferences, it remains unclear whether LLM judges base their decisions on the underlying textual evidence or on superficial, non-evidential cues such as labels, confidence signals, or formatting. In high-stakes applications such as benchmarking, compliance monitoring, and automated reporting, explanations are essential for transparency and auditability; however, if these explanations are generated post hoc to justify decisions influenced by external cues, the evaluation process becomes vulnerable to manipulation and loses its trustworthiness. This problem, which we characterize as rationalization bias, highlights a fundamental gap in current evaluation methodologies: the lack of a principled framework to causally isolate and measure the influence of irrelevant cues on both judgments and explanations. Addressing this gap is essential to ensure that LLM judges produce decisions and rationales that are grounded in evidence, thereby improving the robustness, fairness, and accountability of automated evaluation systems.

\paragraph{Contributions}
\begin{itemize}
\item \textbf{Cue-invariance probing:} We propose a controlled intervention suite (\textsc{Blind}/\textsc{Truth}/\textsc{Flip}/\textsc{Placebo}/\textsc{Reveal-After}) that isolates the causal effect of non-evidential cues on both decisions and explanations while holding texts fixed.
\item \textbf{Anchoring metrics:} We present tie-aware measures for \emph{outcome anchoring} and \emph{rationale anchoring} (label-aligned rhetoric, explanation drift), enabling standardized comparison across judges and prompts.
\item \textbf{Rationalization attacks:} We present a demonstration that verbosity and confidence cues reliably shift outcomes and rationales even for identical summaries, exposing a practical vulnerability in LLM-as-judge pipelines.
\item \textbf{Mitigations.} We evaluate two defenses:
  \begin{itemize}
    \item \emph{Criterion-guided SCoT:} require reasoning along explicit dimensions (e.g., accuracy, completeness, conciseness, fluency) before deciding.
    \item \emph{Proof-Before-Preference (PBP):} \emph{lock} criterion-wise notes with cited spans, then score and rank strictly from the locked evidence, reducing post-hoc rationalization and label anchoring.
  \end{itemize}
\end{itemize}

\section{Related Work}
\subsection{Biases in LLMs as Judges}

A growing line of work has investigated the reliability of LLMs as evaluators, often called the LLM-as-a-judge paradigm. Zheng et al. \cite{10.5555/3666122.3668142} introduced MT-Bench and Chatbot Arena, showing that LLM judges exhibit systematic biases, including position bias, verbosity bias, and self-enhancement bias (favoring their own family of models). Wu \& Aji \cite{wu-aji-2025-style} highlighted a related fluency/style bias, where LLMs prefer eloquent but less accurate answers, rationalizing their decisions by praising surface form.

Broader studies have cataloged biases using benchmark suites. Koo et al. \cite{koo-etal-2024-benchmarking} proposed CoBBLEr, identifying implicit biases (e.g., order, egocentric, salience/length) and induced biases (bandwagon, distractor cues). Ye et al. \cite{ye2025justice} extended this work with Calm, measuring twelve bias types, including authority bias, sentiment bias, and diversity bias, and introducing metrics like robustness rate. Chen et al. \cite{chen-etal-2024-humans} compared human vs. LLM judges, revealing vulnerabilities to misinformation oversight, authority cues, and formatting (“beauty bias”). Lee et al. \cite{hgg} examined judgments of epistemic markers, showing that LLMs penalize expressions of uncertainty while humans do not.

Several works explore adversarial attacks on LLM evaluators. Chen et al. \cite{chen-etal-2024-humans} and Raina et al. \cite{raina-etal-2024-llm} design optimization-based or prompt-based attacks that reliably flip judgments, while Li et al. \cite{li-etal-2024-split} propose bias-mitigation techniques (e.g., randomized ordering). Surveys such as Szymanski et al. \cite{10.1145/3708359.3712091}, Croxford et al. \cite{Croxford2025LLMEval}, and  Thakur et al. (\cite{thakur-etal-2025-judging} emphasize both the promise and fragility of LLM judges: they can align well with human preferences, but remain vulnerable to superficial cues.

\subsection{Rationalization and Explanation Faithfulness in LLMs}

Parallel work examines the faithfulness of model-generated explanations. Turpin et al. \cite{Turpin2023UnfaithfulCoT} demonstrated that LLMs often rely on hidden cues to answer correctly but omit them from their chain-of-thought (CoT), generating post-hoc rationalizations. Chen et al. \cite{Chen2025ReasoningModelsDontAlwaysSay} extended this to state-of-the-art reasoning models, showing low “reveal rates” even when hints clearly influenced answers. Lanham et al. \cite{lanham2023measuringfaithfulnesschainofthoughtreasoning} introduced metrics for CoT faithfulness, such as early-answer and confidence trajectories, while Lewis-Lim et al. \cite{lewislim2025analysingchainthoughtdynamics} analyzed when CoT actively guides reasoning versus narrates predetermined outcomes. Other studies propose methods for improving faithfulness. Chuang et al. \cite{chuang2024faithlmfaithfulexplanationslarge} introduced FaithLM, which ties explanations causally to model outputs by testing against contrary rationales. Li et al. \cite{li-etal-2025-drift} proposed DRiFT, using dual rewards (accuracy + faithfulness) to guide probabilistic inference, improving rationale fidelity. 



\section{Problem Formulation}

Let $D = \{d_1, d_2, \dots, d_N\}$ be a collection of source documents. 
For each document $d \in D$, we generate a set of candidate summaries
\[
S_d = \{s^{\text{ML}}_{1}, \dots, s^{\text{ML}}_{M},\; s^{\text{LLM}}_{1}, \dots, s^{\text{LLM}}_{L}\},
\]
where $\{s^{\text{ML}}_{i}\}$ are produced by traditional machine learning--based extractive systems and $\{s^{\text{LLM}}_{j}\}$ are produced by large language models. An LLM judge $J$ evaluates $S_d$ and returns two outputs:  

\begin{enumerate}
    \item A \textbf{ranking} of candidates, 
    \[
    r_{J,d} \in \mathcal{R}_{M+L},
    \]
    where $\mathcal{R}_{M+L}$ is the set of permutations of the $M+L$ summaries.  

    \item An \textbf{explanation} or rationale,
    \[
    e_{J,d} = f_J(S_d),
    \]
    in free-text form, to justify the ranking.  
\end{enumerate}

\subsection{From Faithfulness to Cue Invariance}

Classically, an explanation $e_{J,d}$ is called \emph{faithful} if it reflects exactly the evidential features $\mathcal{X}_d$ of the fixed texts (source and candidate summaries) that determine the judge’s ranking $r_{J,d}$. A standard formalization is the conditional invariance
\begin{equation}
\Pr\!\big(r_{J,d}\mid \mathcal{X}_d, e_{J,d}\big) \;=\; \Pr\!\big(r_{J,d}\mid \mathcal{X}_d\big),
\label{eq:faithfulness}
\end{equation}
which asserts that, given the evidence $\mathcal{X}_d$, exposing the explanation does not change the distribution of decisions, i.e., the explanation neither adds spurious signal nor reflects hidden, non-evidential influences. In practice, LLM judges can produce \emph{rationalized} explanations: texts that are persuasive to humans yet cite features not causally responsible for the decision. In such cases, \eqref{eq:faithfulness} fails:
\begin{equation}
\Pr\!\big(r_{J,d}\mid \mathcal{X}_d, e_{J,d}\big) \;\neq\; \Pr\!\big(r_{J,d}\mid \mathcal{X}_d\big).
\end{equation}
We refer to the systematic production of such plausible-but-unfaithful rationales as rationalization bias. Directly verifying \eqref{eq:faithfulness} is challenging because we cannot exhaustively observe or control all evidential factors in $\mathcal{X}_d$. We therefore operationalize reliability through \emph{cue invariance}. Let $C$ denote \emph{non-evidential cues} (e.g., labels, badges, stylistic hints). Holding $\mathcal{X}_d$ fixed, a judge is \emph{cue-invariant} if perturbing $C$ leaves both the decision and the explanation (in expectation over measurable properties $\phi$ of the rationale) unchanged:
\begin{align}
\Pr\!\big(r_{J,d}\mid \mathcal{X}_d, C\big) &= \Pr\!\big(r_{J,d}\mid \mathcal{X}_d\big), \label{eq:cue-inv-outcome}\\
\mathbb{E}\!\left[\phi\!\left(e_{J,d}\right)\middle|\mathcal{X}_d, C\right] &= \mathbb{E}\!\left[\phi\!\left(e_{J,d}\right)\middle|\mathcal{X}_d\right]. \label{eq:cue-inv-expl}
\end{align}
Violations of \eqref{eq:cue-inv-outcome} indicate \emph{outcome anchoring}; violations of \eqref{eq:cue-inv-expl} indicate \emph{rationale anchoring} (e.g., increased label-aligned rhetoric or greater explanation drift). In our framework, we intervene on $C$ via controlled conditions (Blind, Truth, Flip, Placebo, Reveal-After) while keeping $\mathcal{X}_d$ fixed, and we quantitatively estimate the resulting shifts in $r_{J,d}$ and $e_{J,d}$. This causal, cue-centric view provides a practical surrogate for faithfulness: when a judge is cue-invariant, its decisions and explanations are robust to non-evidential perturbations, thereby reducing opportunities for post-hoc rationalization.

\section{Methodology}

Our methodology has five components: dataset construction, cue-based causal probing, anchoring metrics, mitigation strategies, and anchoring attacks.

\subsection{Dataset and Notation}
For each $d\in D$, we construct two candidate summaries drawn from traditional extractive systems and LLM. Each summary $s_i$ has observable textual properties (e.g., length, fluency, coverage, factuality), denoted collectively by $\mathcal{X}(s_i)$. We write $\mathcal{X}_d=\{\mathcal{X}(s_i)\}_{i=1}^2$ for the feature set associated with $S_d$.  A judge $J\in\mathcal{J}$ observes $S_d$ under a probe condition $p\in\mathcal{P}$ and produces: (i) a ranking $r^{(p)}_{J,d}\in\mathcal{R}_K$ (for experiments, we consider K=2, i.e one LLM and one traditional ML method), and (ii) a free-text explanation $e^{(p)}_{J,d}=f_J(S_d, p)$. To control trivial position effects, the presentation order of $S_d$ is randomized for each trial.

\subsection{Cue-Based Causal Probing}
We intervene on \emph{non-evidential cues} while holding the texts fixed. Let $C$ denote the cue vector shown with $S_d$ (e.g., source labels such as \textsc{LLM}/\textsc{ML}, or Placebo labels). The ground-truth labels for $S_d$ are $C^\ast$. We define five probe conditions:
\begin{align*}
B &:~ \operatorname{do}(C=\varnothing) && \text{(Blind)}\\
T &:~ \operatorname{do}(C=C^\ast) && \text{(Truth)}\\
F &:~ \operatorname{do}(C=\pi(C^\ast)) && \text{(Flip)}\\
P &:~ \operatorname{do}(C=C^{\text{placebo}}) && \text{(Placebo)}\\
R &:~ \operatorname{do}(C=\pi(C^\ast)) \\
  &\to \operatorname{do}(C=C^\ast) && \text{(Reveal-After)} .
\end{align*}

Here, $B$ hides cues; $T$ reveals correct cues; $F$ reveals inverted cues; $P$ attaches credible but non-informative badges; and $R$ first applies FLIP, then reveals TRUTH.

\paragraph{Cue invariance as our reliability target}
Directly asserting explanation \emph{faithfulness} is intractable without full control of all evidential factors in $\mathcal{X}_d$. We therefore operationalize reliability via \emph{cue invariance}. Holding $\mathcal{X}_d$ fixed, a judge is cue-invariant if perturbing $C$ does not change (a) the decision distribution and (b) measurable properties of the explanation as shown in Eq. \eqref{eq:cue-inv-outcome} and \eqref{eq:cue-inv-expl} . Our probes estimate these effects by contrasting $p\in\{T,F,P,R\}$ against the blind baseline $B$ while keeping $S_d$ (and thus $\mathcal{X}_d$) fixed.

\subsection{Anchoring Metrics}
We quantify how non-evidential cues affect \emph{outcomes} and \emph{explanations} when texts are held fixed. 
Let $\mathcal{O}=\{[1,2],[2,1],\mathrm{Tie}\}$ be the outcome set for a two-candidate comparison.
For probe $p\in\{\mathrm{B},\mathrm{T},\mathrm{F},\mathrm{P},\mathrm{R}\}$, define
\[
p^{(p)}_{J,o} \;=\; \Pr\!\big(r^{(p)}_{J,d}=o\big),\quad o\in\mathcal{O},
\qquad
t^{(p)}_J \;=\; p^{(p)}_{J,\mathrm{Tie}}.
\]
We use $\mathrm{B}$ (Blind) as the reference condition.

\paragraph{Equality Detection Rate (EDR):}
EDR measures whether the judge recognizes that two candidates are effectively equal \emph{when they truly are}.
Let $\mathcal{C}_{\text{eq}}\subseteq D$ denote comparisons labeled as equal (e.g., identical summaries or near-duplicates by a predefined content-equality heuristic).
Then:
\[
\boxed{\ \mathrm{EDR}(J) \;=\; 
\frac{1}{|\mathcal{C}_{\text{eq}}|}
\sum_{d\in\mathcal{C}_{\text{eq}}}
\mathbb{I}\!\big[r^{(\mathrm{B})}_{J,d}=\mathrm{Tie}\big]\ ;}
\]
Here, $\mathbb{I}\{\cdot\}$ is an Indicator function which provides $1$ if the condition is true, $0$ otherwise.
Higher EDR is better (fewer spurious preferences when cues are hidden).

\paragraph{Neutrality Deviation under Blind (tie-aware):}
Among non-ties under Blind, a neutral judge should split $[1,2]$ vs.\ $[2,1]$ at $50{\small\%}$-$50{\small\%}$.
Let
\[
p^{(\mathrm{B})}_{J,12\mid \neg\mathrm{Tie}}
=\frac{p^{(\mathrm{B})}_{J,12}}{p^{(\mathrm{B})}_{J,12}+p^{(\mathrm{B})}_{J,21}},
\qquad
t^{(\mathrm{B})}_J = \Pr(r^{(\mathrm{B})}_{J,d}=\mathrm{Tie}).
\]
Then
\[
\boxed{\ \mathrm{ND}_{\mathrm{B}}(J)
= \bigl|\,2\,p^{(\mathrm{B})}_{J,12\mid \neg\mathrm{Tie}}-1\,\bigr|\cdot \bigl(1-t^{(\mathrm{B})}_J\bigr)\ ;}
\]
when ties are disallowed, $t^{(\mathrm{B})}_J{=}0$ and $\mathrm{ND}_{\mathrm{B}}=|2p^{(\mathrm{B})}_{J,12}-1|$.
Lower is better (more neutral Blind behavior).

\paragraph{Tie-aware, Directional Label Susceptibility (LAO):}
We compare each labeled probe to \emph{the same scheme's} Blind and count only movement \emph{toward} the label-favored decision, while not penalizing movement into $\mathrm{Tie}$.
For $p\in\{\mathrm{T},\mathrm{F}\}$, let the favored non-tie outcome be
\[
\mathrm{fav}_\mathrm{T}=[1,2],\qquad \mathrm{fav}_\mathrm{F}=[2,1],
\]
and $\mathrm{opp}_p$ the other non-tie.
Let
\[
\begin{aligned}
\Delta_{\mathrm{fav}}^{(p)} &= p_{J,\mathrm{fav}_p}^{(p)} - p_{J,\mathrm{fav}_p}^{(\mathrm{B})},\\
\Delta_{\mathrm{opp}}^{(p)} &= p_{J,\mathrm{opp}_p}^{(p)} - p_{J,\mathrm{opp}_p}^{(\mathrm{B})},\\
\Delta_{\mathrm{tie}}^{(p)} &= t_{J}^{(p)} - t_{J}^{(\mathrm{B})}.
\end{aligned}
\]

Decompose positive ``movement-in'' mass:
\[
\begin{aligned}
\mathrm{LDS}^{(p)}&=\max\!\bigl(0,\Delta_{\mathrm{fav}}^{(p)}\bigr),\\
\mathrm{OLS}^{(p)}&=\max\!\bigl(0,\Delta_{\mathrm{opp}}^{(p)}\bigr),\\
\mathrm{TS}^{(p)}&=\max\!\bigl(0,\Delta_{\mathrm{tie}}^{(p)}\bigr).
\end{aligned}
\]

Then
\[
\boxed{%
\mathrm{LAO}(J;p)=
\frac{\mathrm{LDS}^{(p)}}{\mathrm{LDS}^{(p)}+\mathrm{OLS}^{(p)}+\mathrm{TS}^{(p)}+\varepsilon}
}
\]
\vspace{-2pt}
\noindent{\scriptsize $p\in\{\mathrm{T},\mathrm{F}\}$.}

with small $\varepsilon{>}0$ for numerical stability. Lower is better (less label-directed movement). We also report the \emph{absolute} label-directed shift
\[
\boxed{\ \mathrm{LDS}^{(p)}=\max(0,\Delta_{\mathrm{fav}}^{(p)})\ ;\ }
\]
to show magnitude, not only fraction.

\paragraph{Label-Aligned Rationale on Same-Decision (Flip).}
Let $\mathcal{C}_{\text{same}}^{(\mathrm{F})}=\{d\in D:\ r^{(\mathrm{F})}_{J,d}=r^{(\mathrm{B})}_{J,d}\}$ be items whose verdict under \textsc{Flip} matches \textsc{Blind}. 
We quantify label-aligned rhetoric with a hard, temperature-0 embedding scorer $\alpha(e,L^{(\mathrm{F})})\in[0,1]$: 
letting $\mathbf{e}$ be the explanation embedding and $\mathbf{d}_{\mathrm{fav}},\mathbf{d}_{\mathrm{opp}}$ the embeddings of favored/opposite label descriptors under \textsc{Flip}, set
\[
\alpha(e,L^{(\mathrm{F})})=
\begin{cases}
1,& \cos(\mathbf{e},\mathbf{d}_{\mathrm{fav}})>\cos(\mathbf{e},\mathbf{d}_{\mathrm{opp}}),\\
0,& \cos(\mathbf{e},\mathbf{d}_{\mathrm{opp}})>\cos(\mathbf{e},\mathbf{d}_{\mathrm{fav}}),\\
0.5,& \text{otherwise.}
\end{cases}
\]
Then
\[
\boxed{\
\mathrm{LAI}_{\text{text}}\!\mid\!\mathrm{SD}(\mathrm{F})
=
\frac{1}{|\mathcal{C}_{\text{same}}^{(\mathrm{F})}|}
\sum_{d\in\mathcal{C}_{\text{same}}^{(\mathrm{F})}}
\alpha\!\big(e^{(\mathrm{F})}_{J,d},\,L^{(\mathrm{F})}\big)\ ;}
\]
larger values indicate stronger label-aligned rhetoric even when the verdict is unchanged.

\paragraph{Explanation Shift on Same-Decision (Flip).}
Let $\mathcal{C}_{\text{same}}^{(\mathrm{F})}
= \{d \in D : r^{(\mathrm{F})}_{J,d} = r^{(\mathrm{B})}_{J,d}\}$.
We measure how much the \emph{textual explanation itself} changes
when misleading labels are shown, while the verdict remains fixed.

For two explanations $x$ and $y$, we define a normalized cosine
distance:
\[
\Dist(x,y)
= \frac{1 - \cos(x,y)}{2} \in [0,1],
\]
where $0$ indicates identical explanations and $1$ indicates
maximal dissimilarity.

The average explanation drift is
\[
\boxed{
\Delta e \mid \mathrm{SD}(\mathrm{F})
=
\frac{1}{|\mathcal{C}_{\text{same}}^{(\mathrm{F})}|}
\sum_{d \in \mathcal{C}_{\text{same}}^{(\mathrm{F})}}
\Dist\!\big(e^{(\mathrm{F})}_{J,d},\, e^{(\mathrm{B})}_{J,d}\big)
}
\]

and we report a thresholded change rate
\[
\boxed{
\delta^{(\mathrm{F})}_J(\tau)
=
\frac{1}{|\mathcal{C}_{\text{same}}^{(\mathrm{F})}|}
\sum_{d \in \mathcal{C}_{\text{same}}^{(\mathrm{F})}}
\mathbb{I}\!\Big[
\Dist\!\big(e^{(\mathrm{F})}_{J,d}, e^{(\mathrm{B})}_{J,d}\big) > \tau
\Big]
}
\]
with $\tau \in [0,1)$ fixed \emph{a priori}.
If $|\mathcal{C}_{\text{same}}^{(\mathrm{F})}| = 0$, both metrics are
reported as n/a.

\subsection{Mitigation Strategies}
We study two complementary defenses:

\textbf{Structured Chain-of-Thought (SCoT):}
Judges are instructed to evaluate along predefined criteria 
$\mathcal{C}=\{\text{accuracy},\text{completeness},\text{conciseness},\text{fluency}\}$.
For each $s_i$, the judge outputs criterion-specific scores (and optionally cited spans), and we aggregate with nonnegative weights $\{w_c\}_{c\in\mathcal{C}}$ (typically $\sum_c w_c=1$):
\[
r^{(p)}_{J,d}=\mathrm{rank}\Big(\big\{\sum_{c\in\mathcal{C}} w_c \cdot \text{score}_c(s_i)\big\}_{i=1}^K\Big).
\]
Structuring the rationale can reduce scope for free-form post-hoc justifications.

\textbf{Proof-Before-Preference (PBP):}
PBP enforces \emph{evidence lock} before any preference is stated. 
Given candidates $\{s_i\}_{i=1}^K$ and criteria $\mathcal{C}$, Turn~1 collects for each $(i,c)$ a short note $n_{i,c}$ with cited spans from the source; these notes are then locked, $\mathrm{lock}(\{n_{i,c}\})$, prohibiting edits.
Turn~2 assigns scores using only the locked notes:
\[
\text{score}_c(s_i) \;=\; f_c\!\big(\mathrm{lock}(\{n_{i,c}\})\big).
\]
Turn~3 aggregates scores into a final ranking,
\[
r^{(p)}_{J,d} \;=\; \mathrm{rank}\Big(\big\{\sum_{c\in\mathcal{C}} w_c \cdot \text{score}_c(s_i)\big\}_{i=1}^K\Big),
\]
and the narrative justification must reference the locked evidence.
By forcing \emph{evidence before preference}, PBP reduces post-hoc rationalization and label anchoring.

\subsection{Anchoring Attacks}
To further stress-test robustness, we apply semantic-preserving style transformations $\mathcal{A}:S_d\mapsto S_d'$ that inject known cues.

\noindent \textbf{Verbosity Attack:} $\mathcal{A}_{\text{verb}}$ appends redundant but content-preserving text to target summaries, probing whether judges reward length and then rationalize it as ``detail.''

\noindent \textbf{Confidence Attack:} $\mathcal{A}_{\text{conf}}$ rewrites tone to be more assertive without altering factual content, probing whether judges reward certainty and rationalize it as ``precision.''

\section{Experimental Setup}

\subsection{Dataset Construction}
Reliable bias testing requires evaluation data that is (i) \emph{unseen} by the judged models and (ii) \emph{decontaminated} from common pretraining corpora. Widely used benchmarks such as CNN/DailyMail \cite{hermann2015teaching} and XSum \cite{narayan2018don} are likely present, at least in part, in LLM training data, risking confounds from memorization or prior exposure. We therefore curate a new corpus of $N{=}1{,}000$ \emph{summaries} drawn from publicly available sources spanning business, literature, mathematics, science, and technology to ensure topical diversity. Each document is between 500 and 1{,}200 tokens. 

For each document $d$, we create a comparison set $S_d=\{s_1,\dots,s_K\}$ that mixes traditional extractive systems, \textsc{TextRank}, \textsc{LexRank}, \textsc{KL-Sum}, \textsc{SumBasic}, and LLM summaries (instruction-following prompts with fixed decoding parameters). In initial experiments, LLM judges frequently preferred LLM outputs over extractive baselines; importantly, these preferences often reflected \emph{genuine quality gains} (e.g., higher coverage/fluency) rather than mere self-favoring (refer to Table \ref{tml} in Appendix). To \emph{isolate} cue effects from true quality differences, we add two controlled subsets that hold content fixed while perturbing labels:

\paragraph{Equal-content pair subset ($\mathcal{C}_{\text{eq-pair}}$)}
For each $d$, the \emph{same} LLM produces two paraphrases $\tilde{s}_1,\tilde{s}_2$ that preserve semantics but vary superficially. We then vary cues $C$ (e.g., label one as \textsc{LLM} and the other as \textsc{TradML}) across the T/F/P probes, keeping texts fixed. Any movement in outcomes or explanations on $\mathcal{C}_{\text{eq-pair}}$ quantifies \emph{pure} cue anchoring. Most of the experiments are conducted using this approach (except the ones which are explicitly mentioned).

\paragraph{Single-summary relabel subset ($\mathcal{C}_{\text{single}}$)}
We also present the \emph{identical} summary $s$ under different labels across probes while holding the opponent fixed. Because $\mathcal{X}(s)$ is constant, any change in ranking or rationale must be driven by cues $C$, not content refer to Table \ref{tab:identical} in Appendix.

\subsection{Judges}
We employ five open-weight LLMs, Gemma-2-9B, Llama-3.1-8B, Mistral-7B, Qwen2.5-7B, and Zephyr-7B, as \emph{judges}, executed from downloaded checkpoints on our local infrastructure; no third-party APIs were used.  For each document $d$, each judge $J$ is shown the full candidate set $S_d$ under probe condition $p\in\mathcal{P}$ and produces (i) a complete ranking $r^{(p)}{J,d}\in\mathcal{R}{2}$ and (ii) a free-form explanation $e^{(p)}_{J,d}$. Prompts are standardized across models; only the probe condition varies. All judge generations use temperature 
0 to ensure deterministic outputs and reproducibility.
\begin{figure}[h]
\centering
		\subfigure[Equality Detection Rate.]{
		\includegraphics[width=0.42\linewidth]{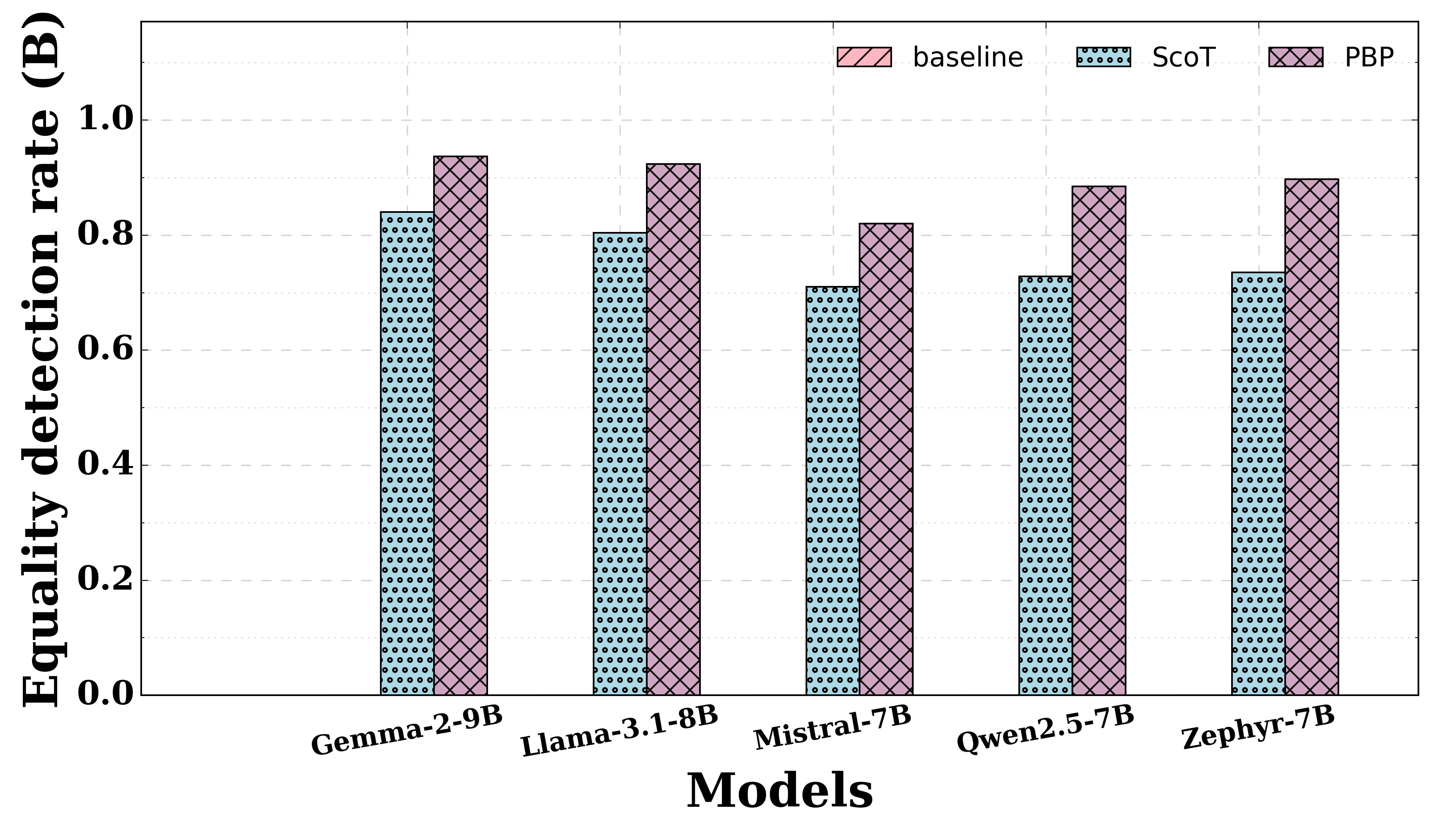}
		\label{edr}	
	}
		\subfigure[Neutrality Deviation.]{
		\includegraphics[width=0.42\linewidth]{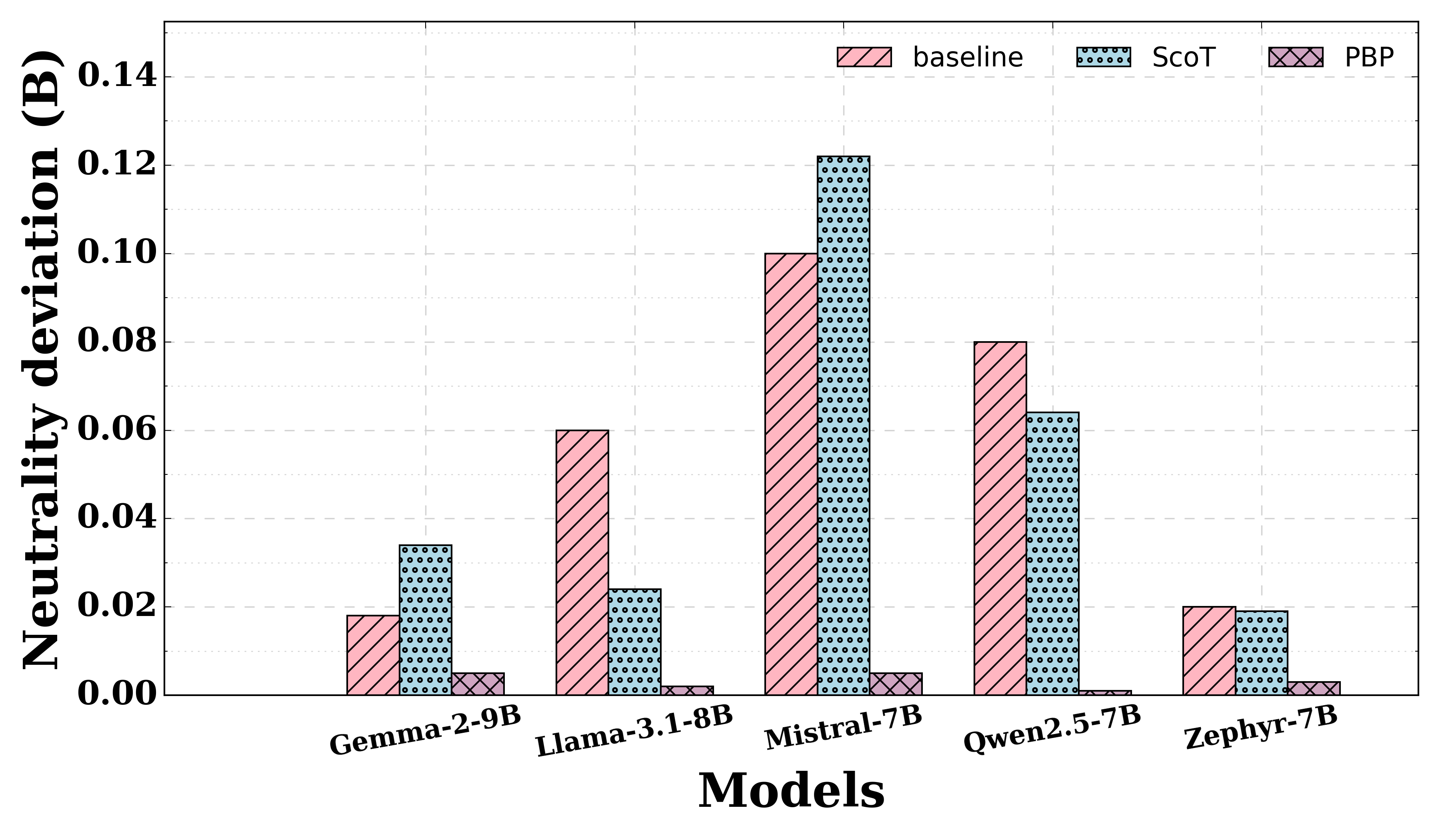}
		\label{nd}	
	}
	\caption{Blind-Condition Behavior of Different Judges.}
	\label{Fig 1}
\end{figure}
\begin{figure}
    \centering
    \includegraphics[width=0.4\textwidth]{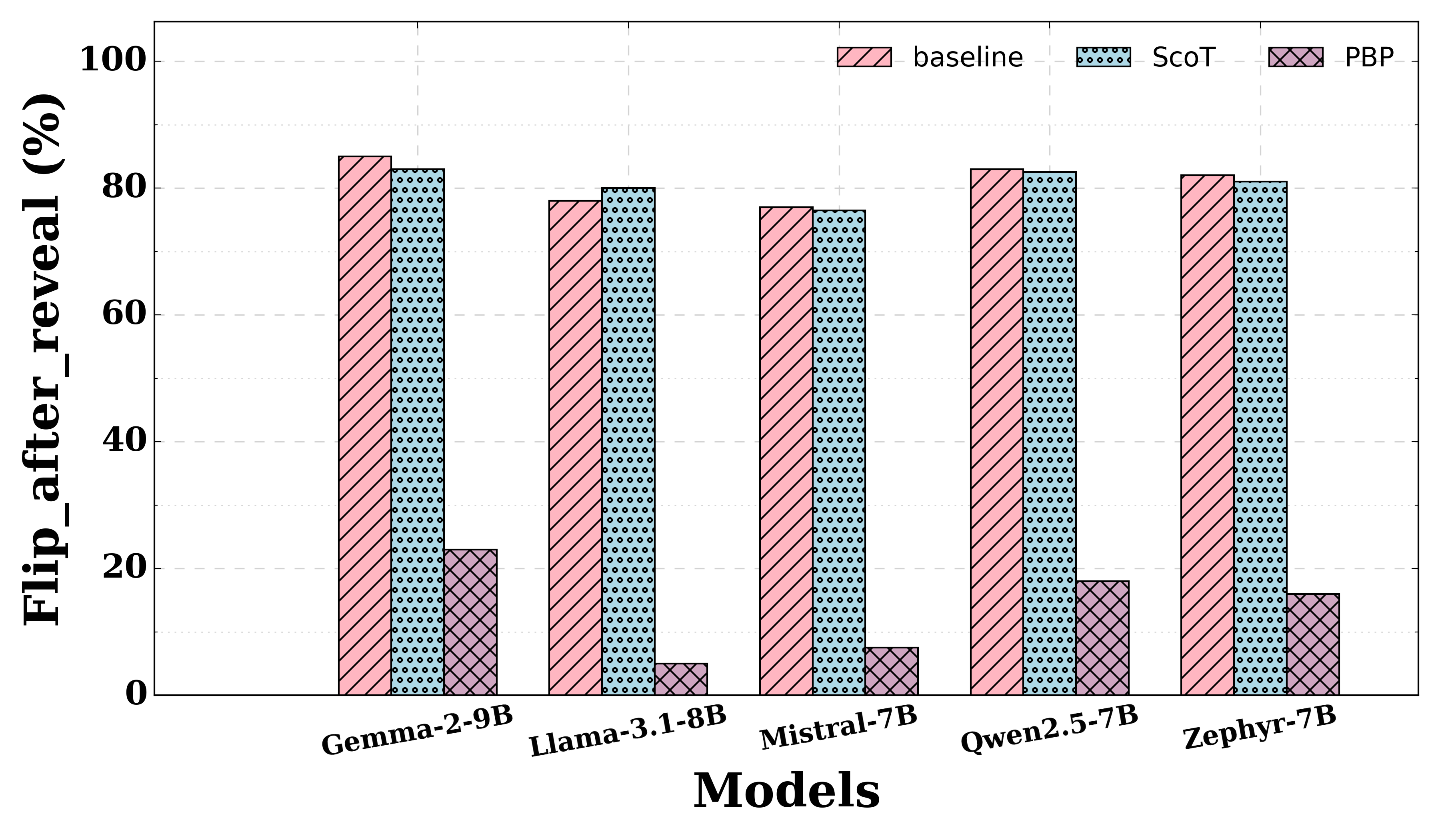}
    \caption{Revision Susceptibility after Label Reveal}
    \label{Fig 8}
\end{figure}
\begin{figure*}[h]
\centering
		\subfigure[Label–Anchoring in Outcomes for True Labels.]{
		\includegraphics[width=0.31\linewidth]{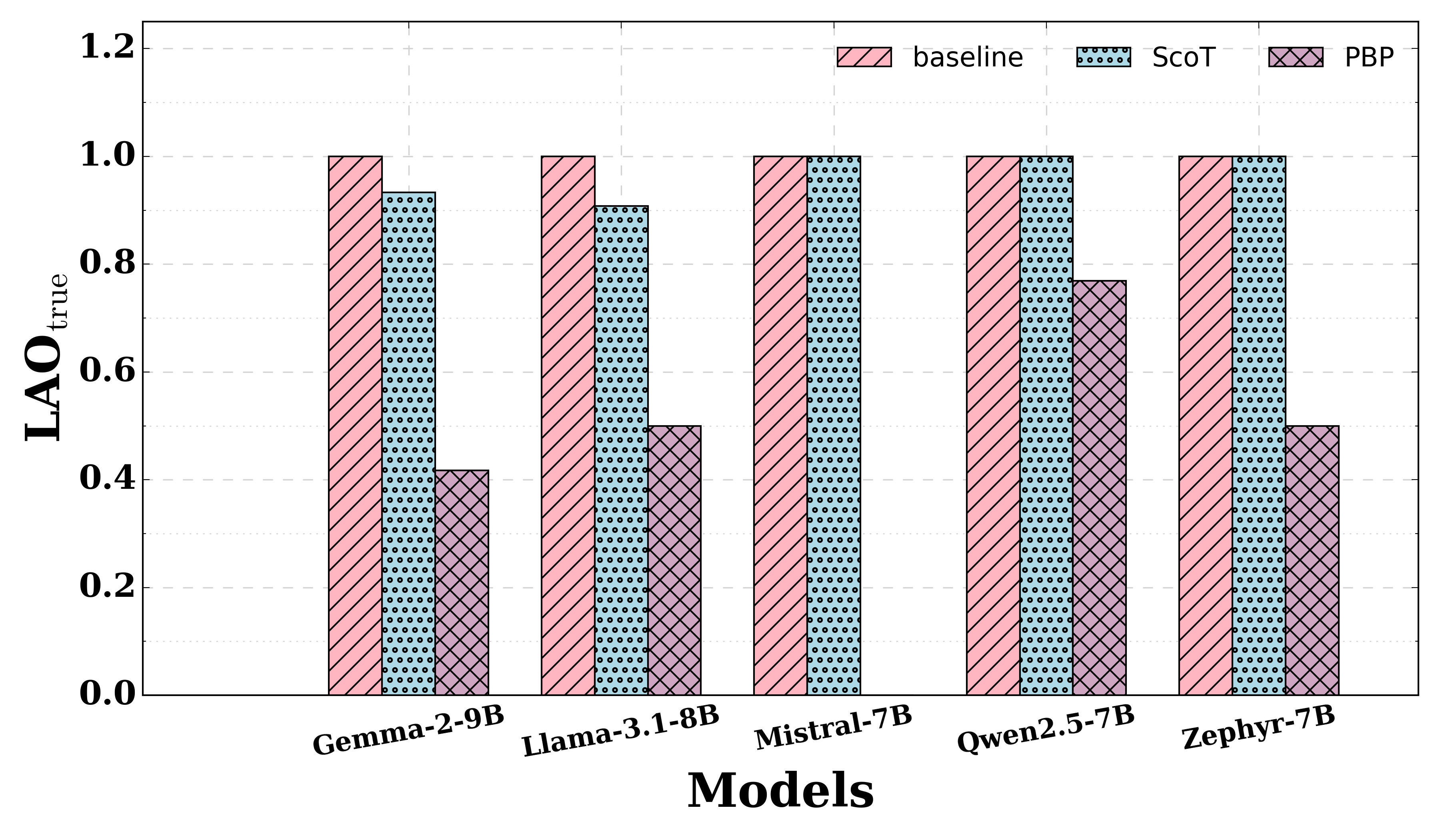}
		\label{laot}	
	}
		\subfigure[Label–Anchoring in Outcomes for Flip Labels.]{
		\includegraphics[width=0.31\linewidth]{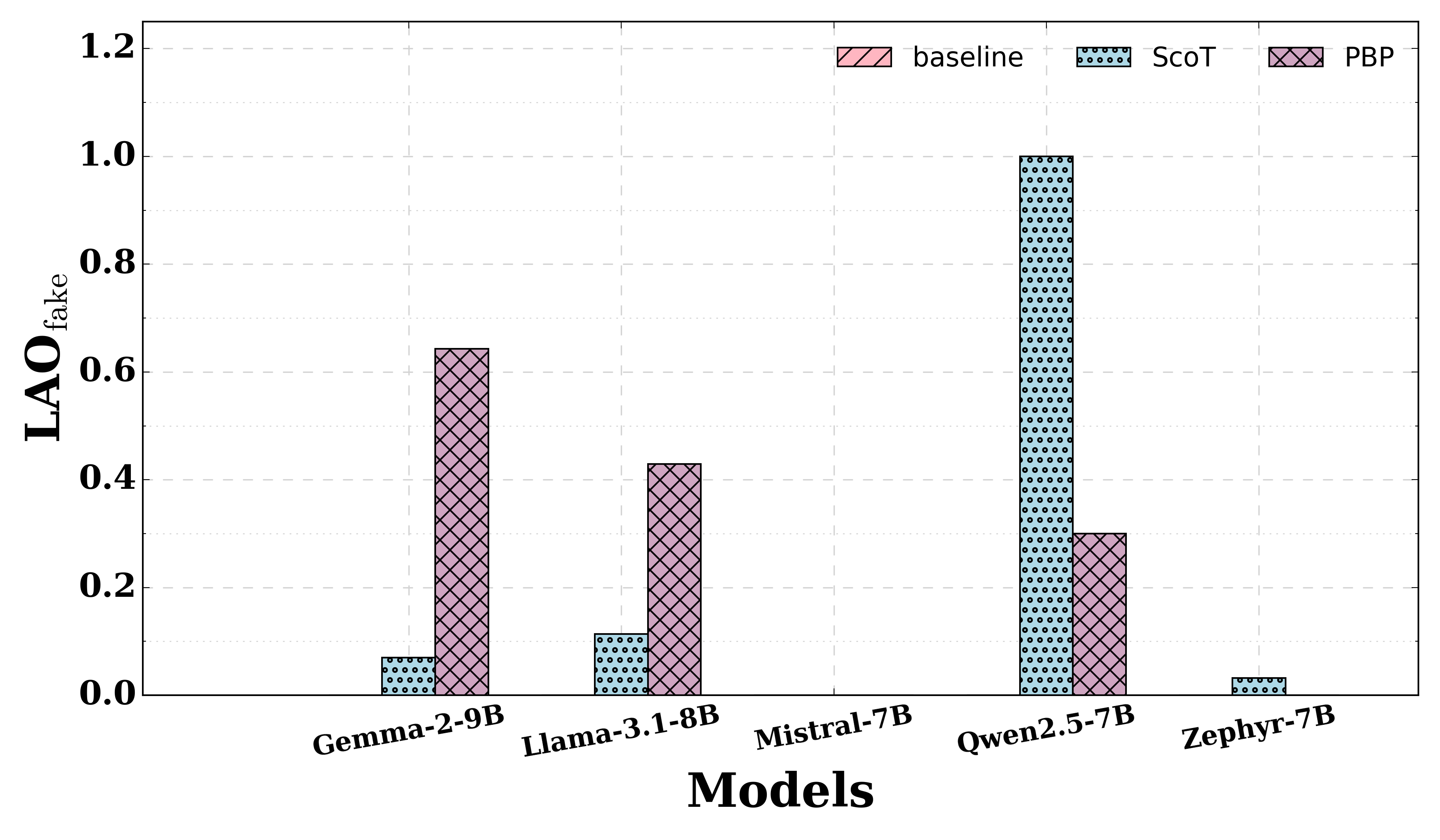}
		\label{laof}	
	}
    \subfigure[Label–Anchoring in Outcomes for Placebo Labels.]{
		\includegraphics[width=0.31\linewidth]{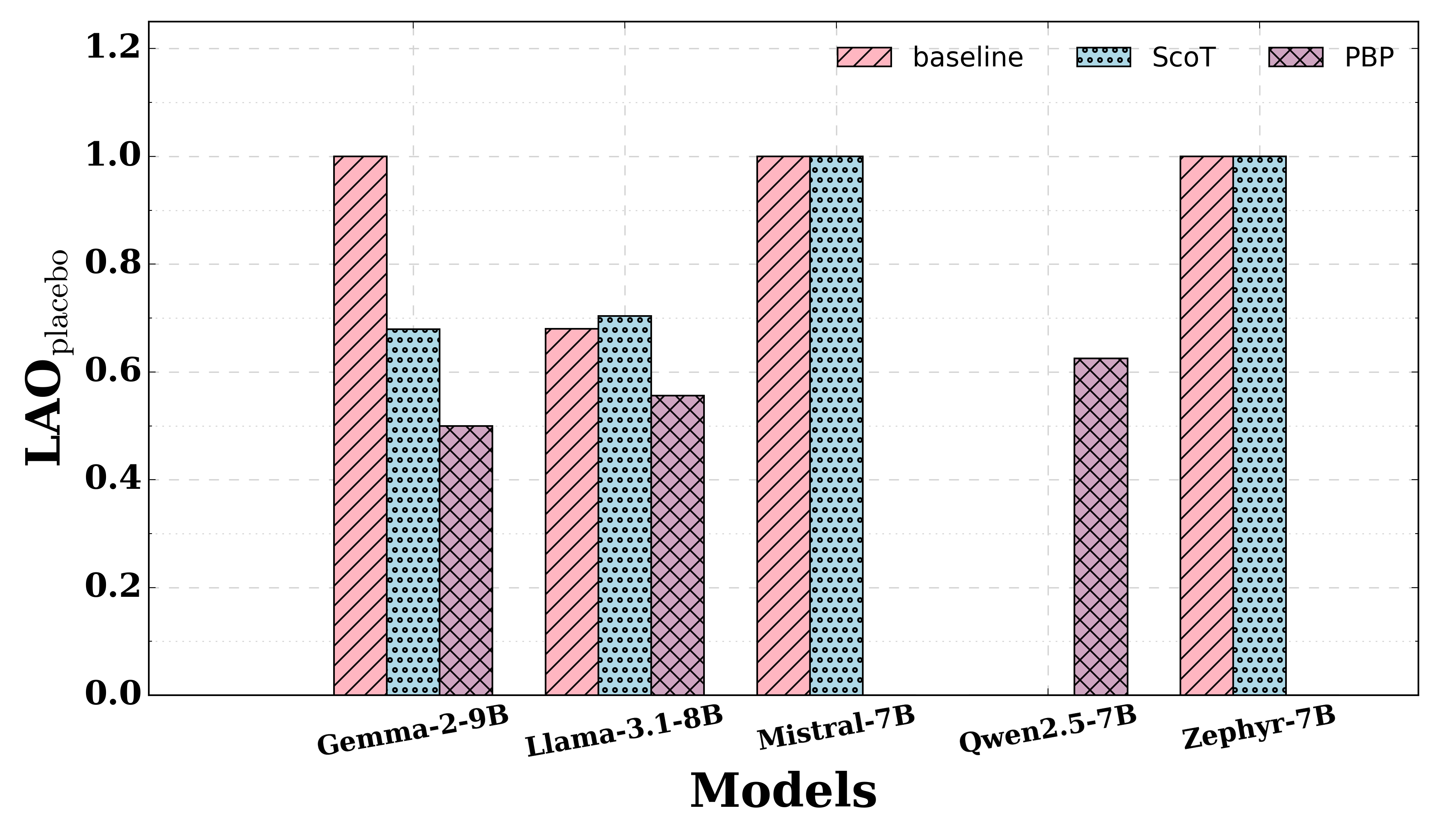}
		\label{laop}	
	}
	
	\caption{Label–Anchoring in Outcomes for Different Judges.}
	\label{Fig 2}
\end{figure*}

\subsection{Prompt Examples}

We document the prompt templates used for \emph{generation} and \emph{judging}. Our goals are (i) \textbf{reproducibility} via frozen, model-agnostic instructions; (ii) \textbf{comparability} across protocols by holding wording constant except for the protocol-specific constraints; and (iii) \textbf{diagnosability} of bias via controlled label probes (\textsc{Blind}, \textsc{True}, \textsc{FLIP}, \textsc{Placebo}, \textsc{Reveal-after}) and style attacks (\textsc{Verbosity}, \textsc{Confidence}). All judges emit structured verdict \& rationale, and all templates forbid external knowledge to isolate effects of labels/cues rather than content leakage. The three protocols differ only in \emph{how} evidence is elicited: \textbf{Baseline} gives a direct verdict with a brief post-hoc explanation; \textbf{SCoT} requires criterion-guided notes and scores (accuracy, completeness, conciseness, fluency) prior to the verdict; and \textbf{PBP} enforces an \emph{evidence lock} (Explain$\rightarrow$Score$\rightarrow$Rank), so decisions and rationales must derive from quoted spans captured before scoring. Placeholders in \texttt{\{BRACES\}} are programmatically filled at runtime and hyperparameters (e.g., temperature, max tokens, seeds) are fixed across conditions.

\begin{promptbox}[title={Summary Generation (LLM author)}]
You are a careful writer. Given a document, produce a concise, faithful
summary of 500 words.

\textbf{Document}: \{\texttt{DOC\_TEXT}\}

\textbf{Output}: \emph{The summary text}.
\end{promptbox}

\begin{promptbox}[title={Baseline Judge (direct verdict, brief rationale)}]
You are a careful judge. Two summaries describe the same document.
Decide which summary is \textbf{better overall} and briefly explain why.

\textbf{Document}: \{\texttt{DOC\_TEXT}\}

\textbf{Summary 1 (label = \{\texttt{LABEL\_1}\})}: \{\texttt{SUM1}\}

\textbf{Summary 2 (label = \{\texttt{LABEL\_2}\})}: \{\texttt{SUM2}\}

\textbf{Output}:
\{\{
 "verdict": "[1,2]" \textbar\ "[2,1]" \textbar\ "Tie",
 "explanation": "2--4 sentences in plain English"
\}\}

\end{promptbox}

\begin{promptbox}[title={SCoT Judge (criterion–guided decision)}]
You are a judge. Evaluate along the following criteria:
\textbf{accuracy}, \textbf{completeness}, \textbf{conciseness}, \textbf{fluency}.
Decide which summary is \textbf{better overall} and briefly explain why.

\textbf{Document}: \{\texttt{DOC\_TEXT}\}

\textbf{Summary 1 (label = \{\texttt{LABEL\_1}\})}: \{\texttt{SUM1}\}

\textbf{Summary 2 (label = \{\texttt{LABEL\_2}\})}: \{\texttt{SUM2}\}

\textbf{Output}:
\{\{
 "verdict": "[1,2]" \textbar\ "[2,1]" \textbar\ "Tie",
 "explanation": "2--4 sentences in plain English"
\}\}
\end{promptbox}

\begin{promptbox}[title={PBP Judge: \emph{Lock Evidence} (no scores, no verdict)}]
You are an impartial judge. First, write criterion-wise notes for \textbf{accuracy},
\textbf{completeness}, \textbf{conciseness}, \textbf{fluency}. For every note, quote
or reference the exact span(s) from the \textbf{Document} or \textbf{Summaries}
that support it. \textbf{Do not} provide a verdict. \emph{Score} using locked notes only
Using \textbf{only} the \texttt{locked\_notes} above, assign 0--5 scores for each
criterion and summary. Do \emph{not} add new evidence. Aggregate the criterion scores (equal weights unless specified) and produce the
overall verdict and a one–sentence justification that references the locked notes.

\textbf{Document}: \{\texttt{DOC\_TEXT}\}

\textbf{Summary 1 (label = \{\texttt{LABEL\_1}\})}: \{\texttt{SUM1}\}

\textbf{Summary 2 (label = \{\texttt{LABEL\_2}\})}: \{\texttt{SUM2}\}

\textbf{Output}:
\{\{
 "verdict": "[1,2]" \textbar\ "[2,1]" \textbar\ "Tie",
 "explanation": "reference locked evidence only"
\}\}
\end{promptbox}

\subsection{Results}

\subsubsection{Blind-Condition Behavior: Equality Detection and Neutrality}

Under the Blind condition (B), we assess two complementary properties of judge behavior: the \emph{Equality Detection Rate} $\mathrm{EDR}_B$ (refer Fig. \ref{edr}), the fraction of items marked \emph{Tie/No selection}, and the \emph{Neutrality Deviation} $\mathrm{ND}_B$ (refer Fig. \ref{nd}), the tie-adjusted deviation from a 50--50 split between $[1,2]$ and $[2,1]$ (lower is better). Taken together, these metrics reveal a consistent pattern across models: \textbf{PBP} exhibits the most faithful Blind behavior, \textbf{SCoT} is intermediate, and the \textbf{Baseline} is worst. Specifically, $\mathrm{EDR}_B$ is essentially zero for Baseline (no abstention), rises for SCoT ($\approx 0.71$--$0.84$), and is highest for PBP ($\approx 0.82$--$0.94$), indicating that PBP most often recognizes near-equivalent summaries. The same ordering holds in reverse for $\mathrm{ND}_B$: Baseline shows the largest deviations (e.g., Gemma $0.018$, Llama $0.060$, Mistral $0.100$, Qwen $0.080$, Zephyr $0.020$), SCoT is modestly skewed ($\approx 0.019$--$0.122$), and PBP is closest to neutral ($\approx 0.001$--$0.005$). For instance, \emph{Gemma-2-9B} moves from $\mathrm{EDR}_B \approx 0$ (Baseline) to $\sim 0.84$ (SCoT) and \textbf{0.937} (PBP), while $\mathrm{ND}_B$ drops from $0.018$ (Baseline) to \textbf{0.005} (PBP); \emph{Llama-3.1-8B} shows a similar reduction in $\mathrm{ND}_B$ from $0.060$ (Baseline) to  \textbf{0.002} (PBP). 

\begin{figure}[h]
\centering
		\subfigure[Explanation drift under Same–Decision.]{
		\includegraphics[width=0.42\linewidth]{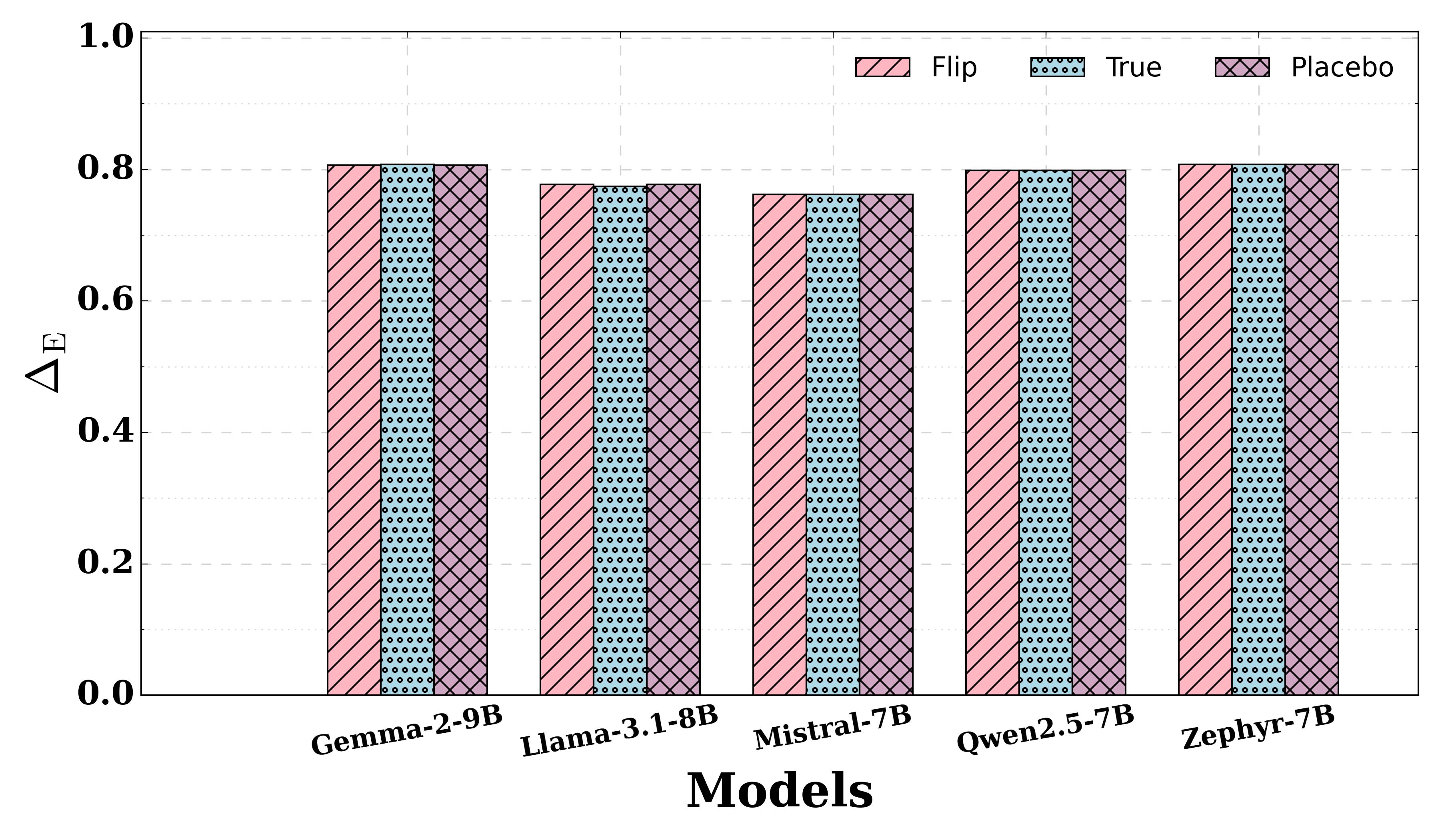}
		\label{delta}	
	}
		\subfigure[Label-aligned explanation under Same–Decision.]{
		\includegraphics[width=0.42\linewidth]{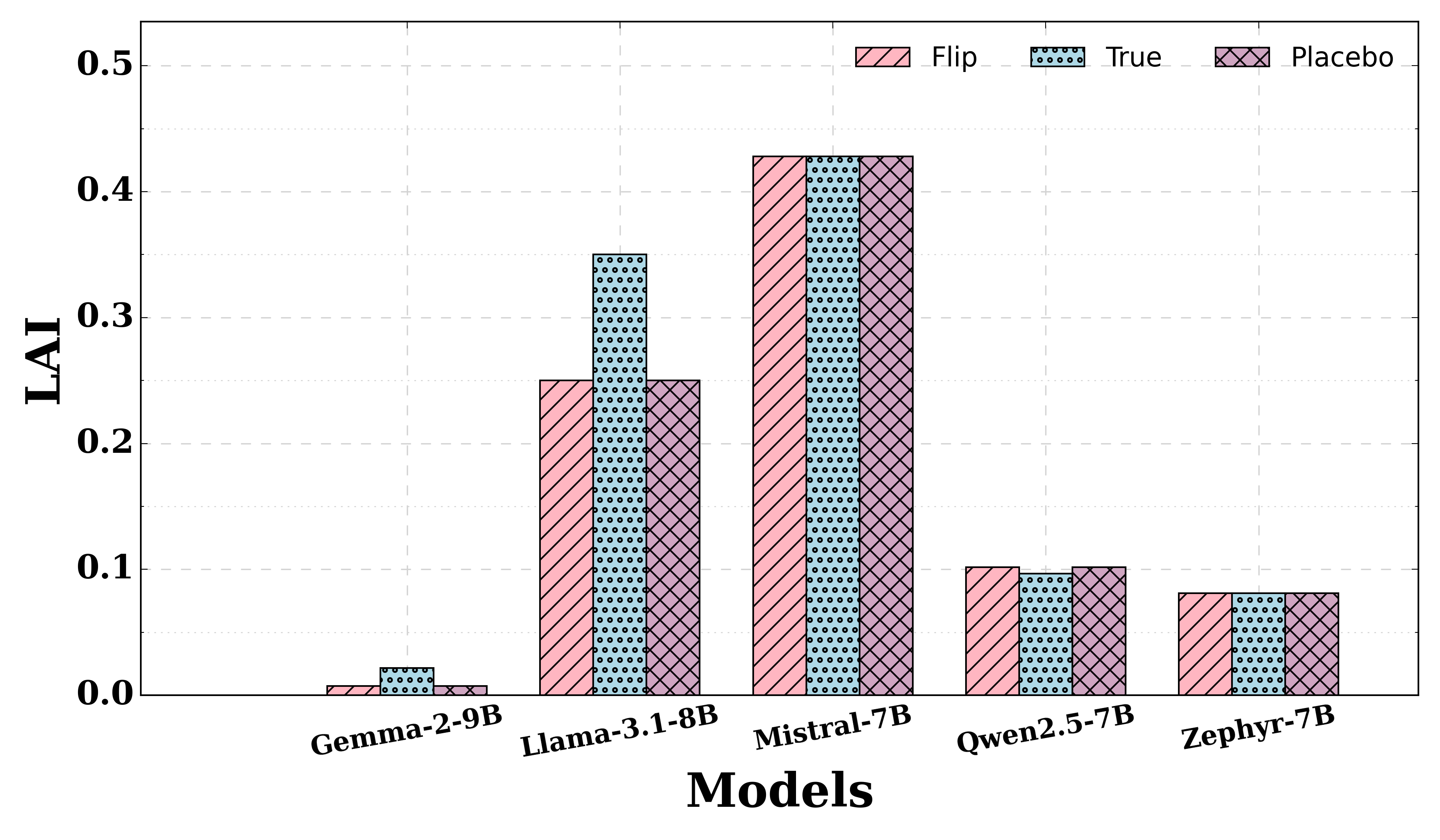}
		\label{lai}	
	}
\caption{Explanation rationalization with the verdict held constant.}
\label{Fig 3}
\end{figure}

\subsection{Label–Anchoring in Outcomes (LAO):}
We quantify label susceptibility using the tie-aware Label--Anchoring in Outcomes (LAO) metric. It captures the fraction of total outcome movement that aligns with the presented label or cue, with higher values indicating stronger label-directed anchoring. Fig. \ref{Fig 2} presents LAO values under three probe conditions: True labels, Flip labels, and Placebo labels. Under True labels (refer Fig.~\ref{laot}), the Baseline protocol exhibits maximal anchoring (LAO = 1.00) across all models. SCoT also shows high anchoring, with LAO values ranging from approximately 0.91 to 1.00. In contrast, PBP consistently exhibits lower LAO values across models (approximately 0.00--0.77), indicating reduced alignment between outcomes and label cues when evidence locking is enforced.

Under Flip labels (refer Fig.~\ref{laof}), where misleading labels are presented, the Baseline protocol shows negligible anchoring (LAO $\approx$ 0), indicating minimal movement toward incorrect label-favored outcomes. SCoT exhibits selective anchoring, with some models showing low LAO and others showing elevated anchoring, notably Qwen2.5-7B approaching LAO $\approx$ 1.00. PBP generally maintains lower anchoring than SCoT, with most values near zero and moderate anchoring observed in only a subset of models.

Under Placebo labels (refer Fig.~\ref{laop}), which introduce non-informative but credible cues, the Baseline protocol again shows strong anchoring in several models (LAO $\approx$ 1.00 in Gemma-2-9B, Mistral-7B, and Zephyr-7B). SCoT also exhibits elevated anchoring across models, with values ranging from approximately 0.68 to 1.00. In contrast, PBP maintains lower LAO values overall, although moderate anchoring is observed in some models (approximately 0.50--0.63), while remaining near zero in others.

Overall, these results show that the Baseline protocol is highly sensitive to label and cue signals, SCoT reduces anchoring under some conditions but remains susceptible when cues are present, and PBP consistently exhibits lower anchoring across probe conditions.

\begin{figure*}[h]
\centering
		\subfigure[Variation in LAO.]{
		\includegraphics[width=0.22\linewidth]{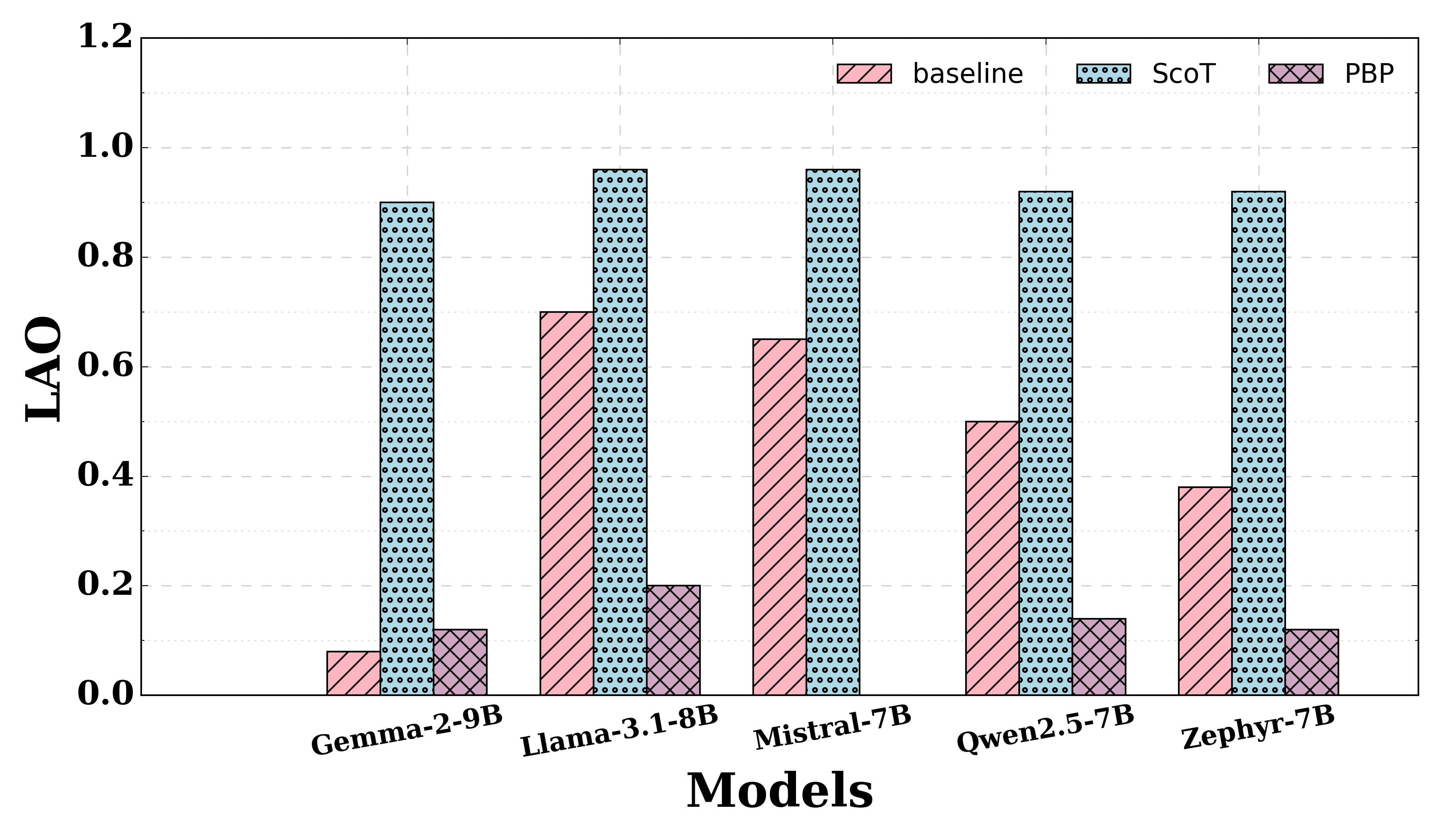}
		\label{f1}	
	}
		\subfigure[Variation in LDS.]{
		\includegraphics[width=0.22\linewidth]{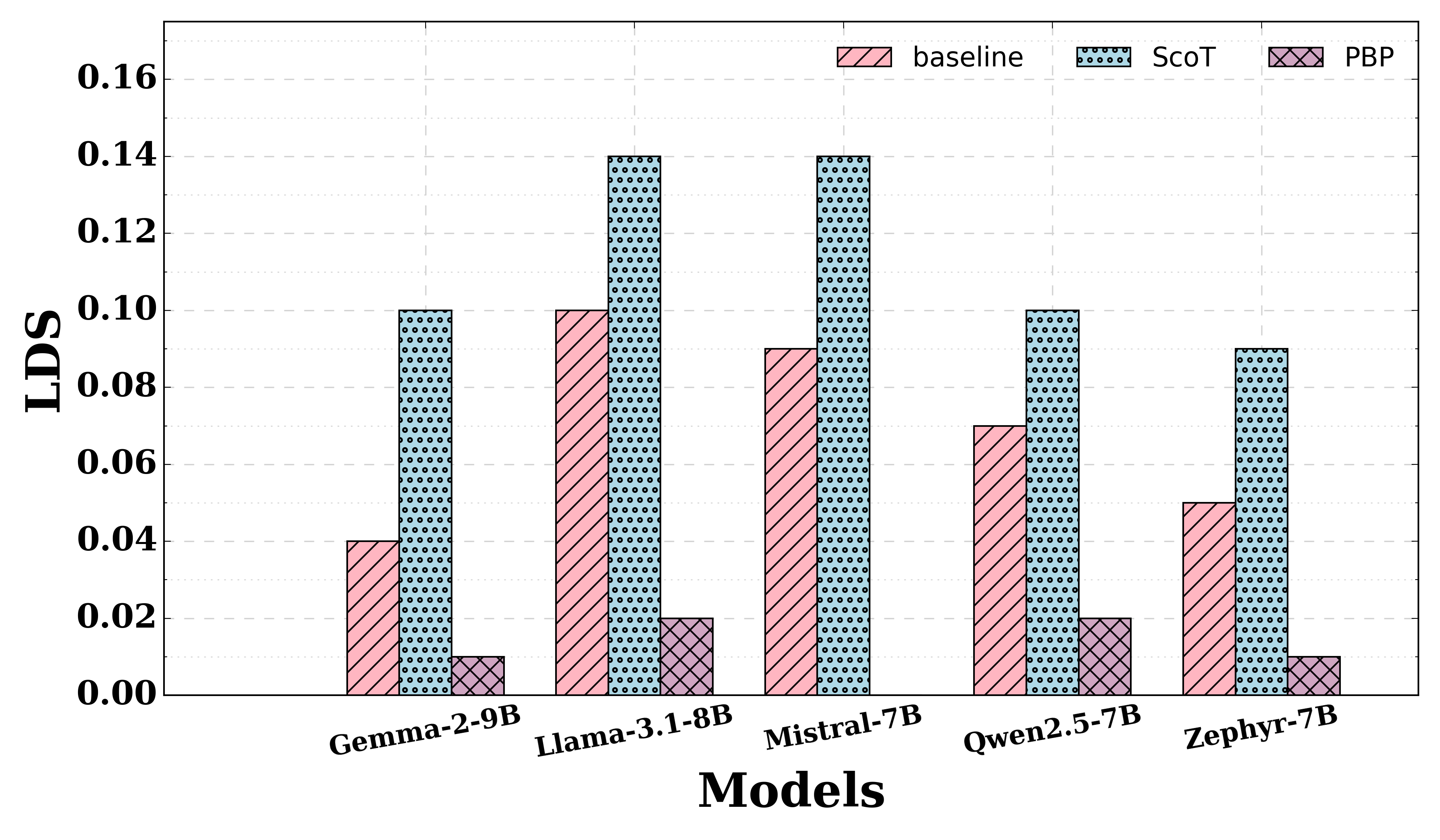}
		\label{f2}	
	}
    \subfigure[Variation in Explanation drift under Same–Decision.]{
		\includegraphics[width=0.22\linewidth]{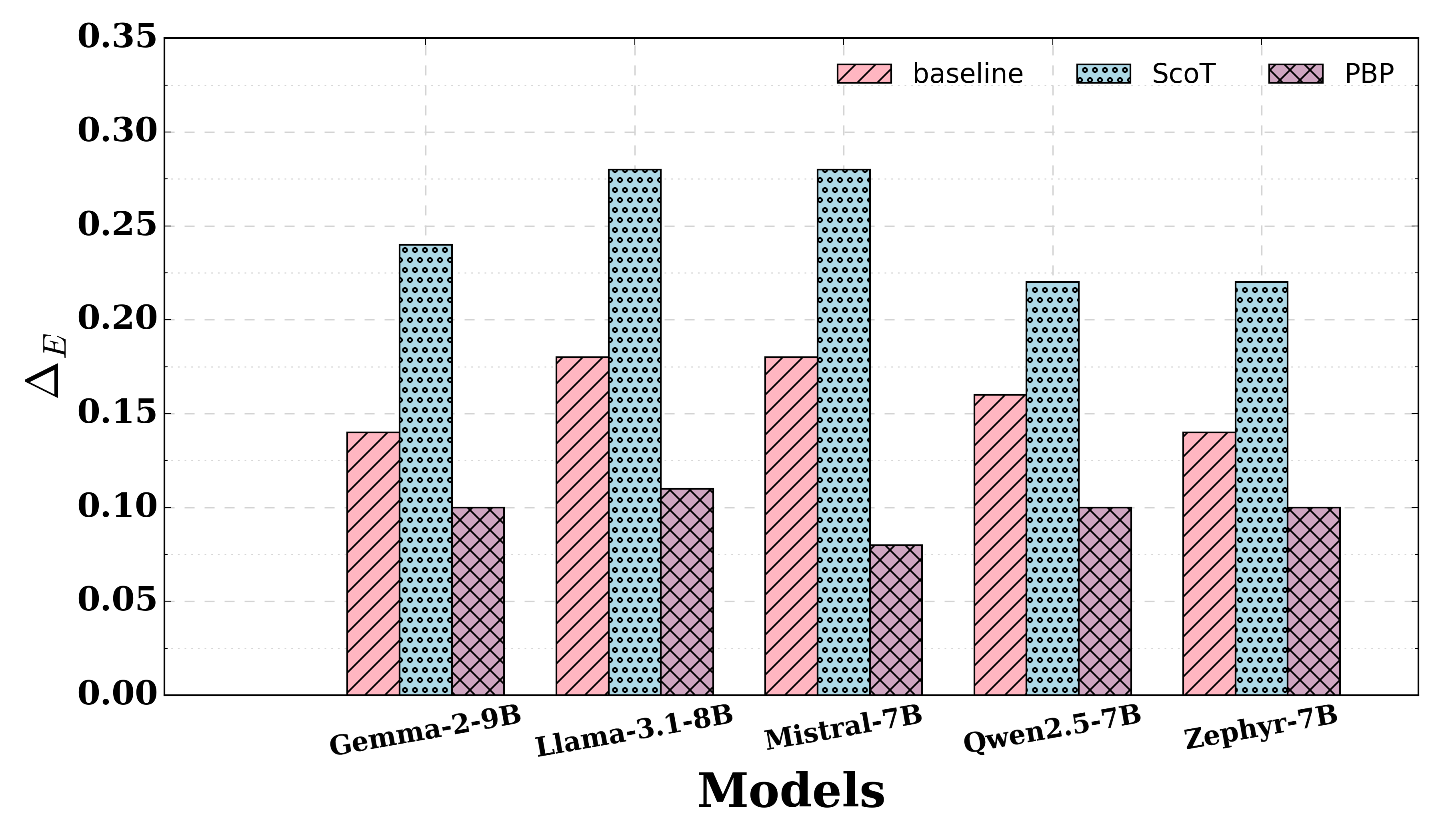}
		\label{f3}	
	}
    \subfigure[Variation in Cue.]{
		\includegraphics[width=0.22\linewidth]{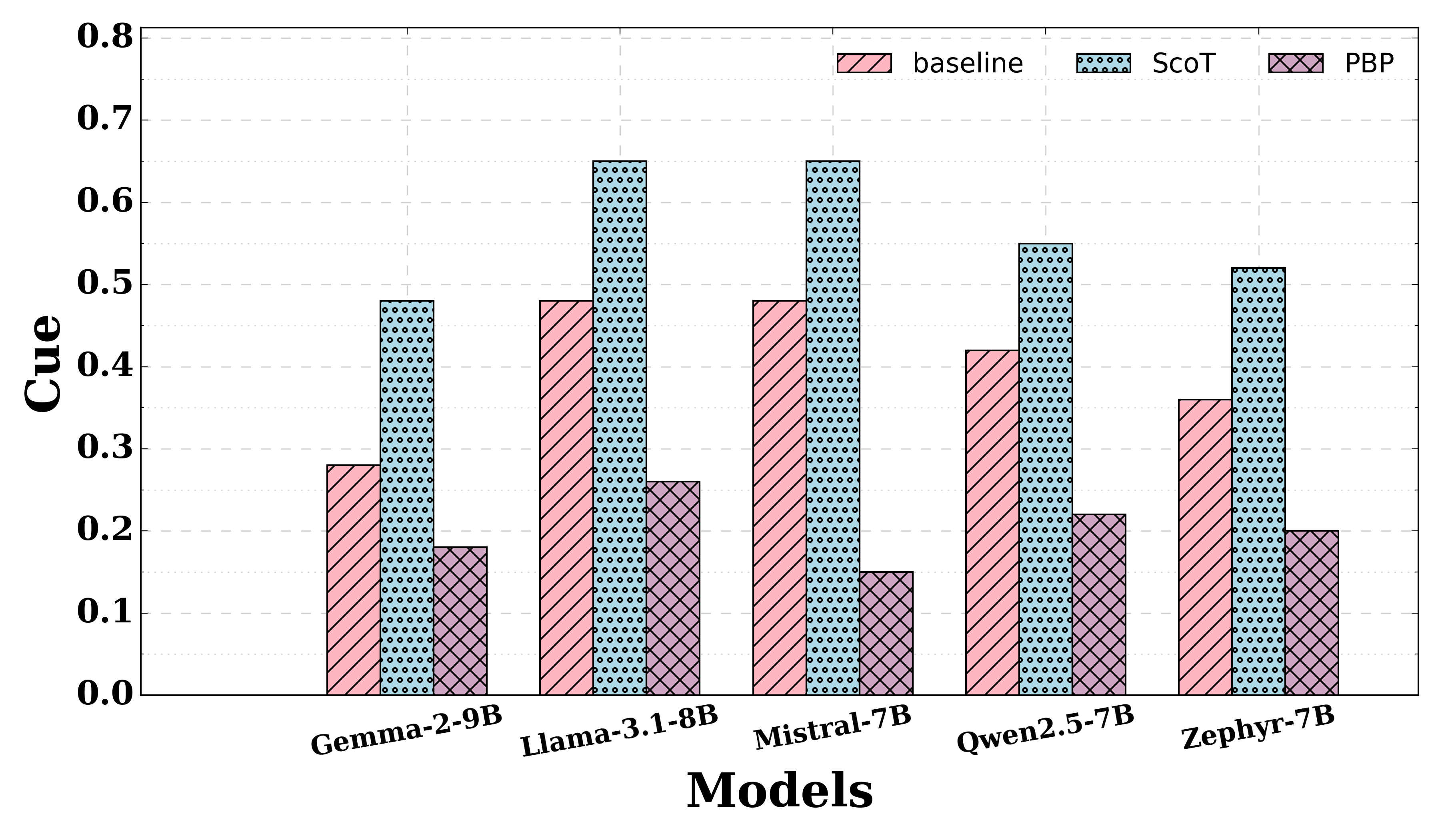}
		\label{f4}	
	}
	\caption{Variation in Different Parameters under Verbosity attack.}
	\label{Fig 4}
\end{figure*}

\begin{figure*}[h]
\centering
		\subfigure[Variation in LAO.]{
		\includegraphics[width=0.22\linewidth]{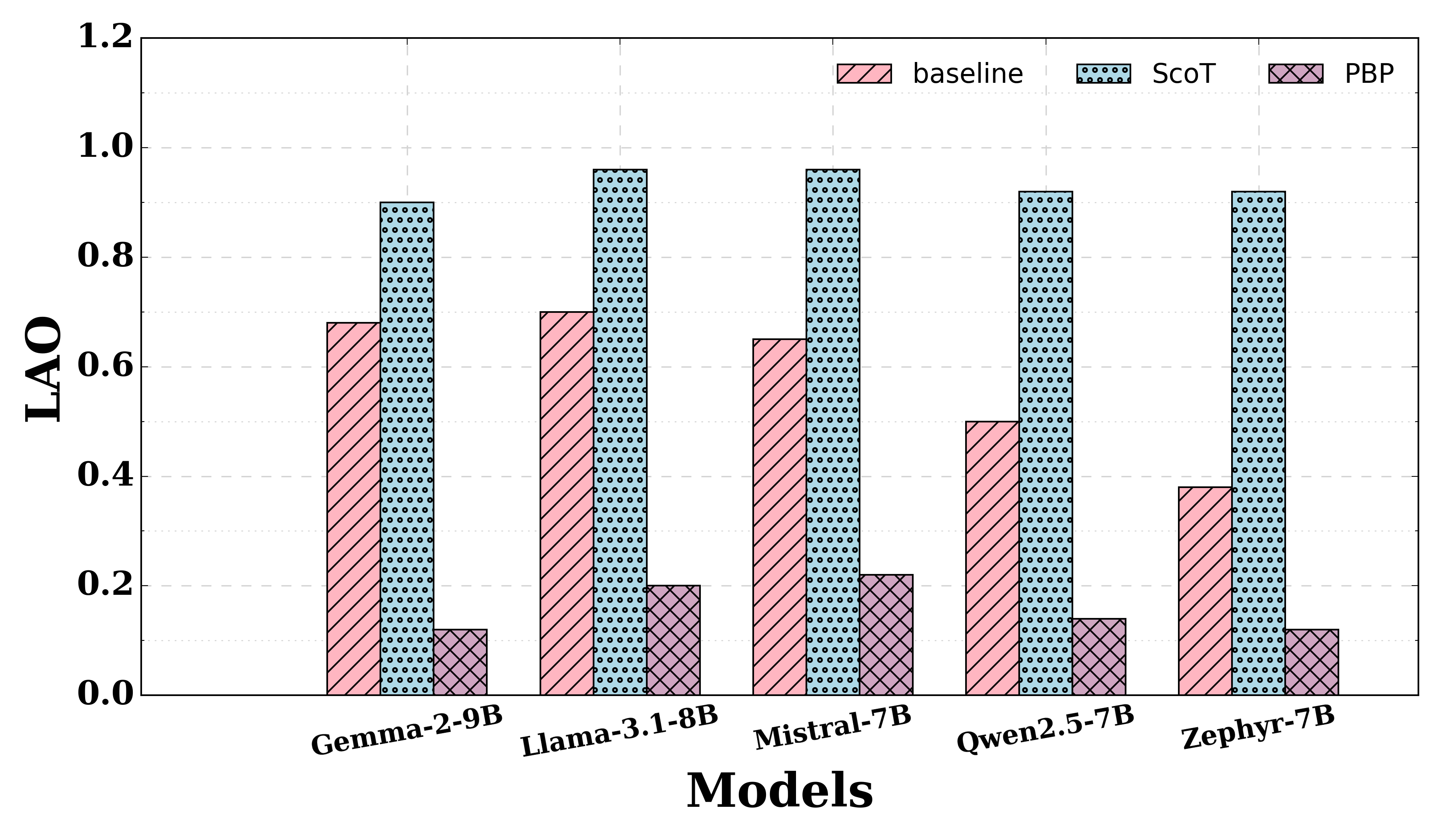}
		\label{f5}	
	}
		\subfigure[Variation in LDS.]{
		\includegraphics[width=0.22\linewidth]{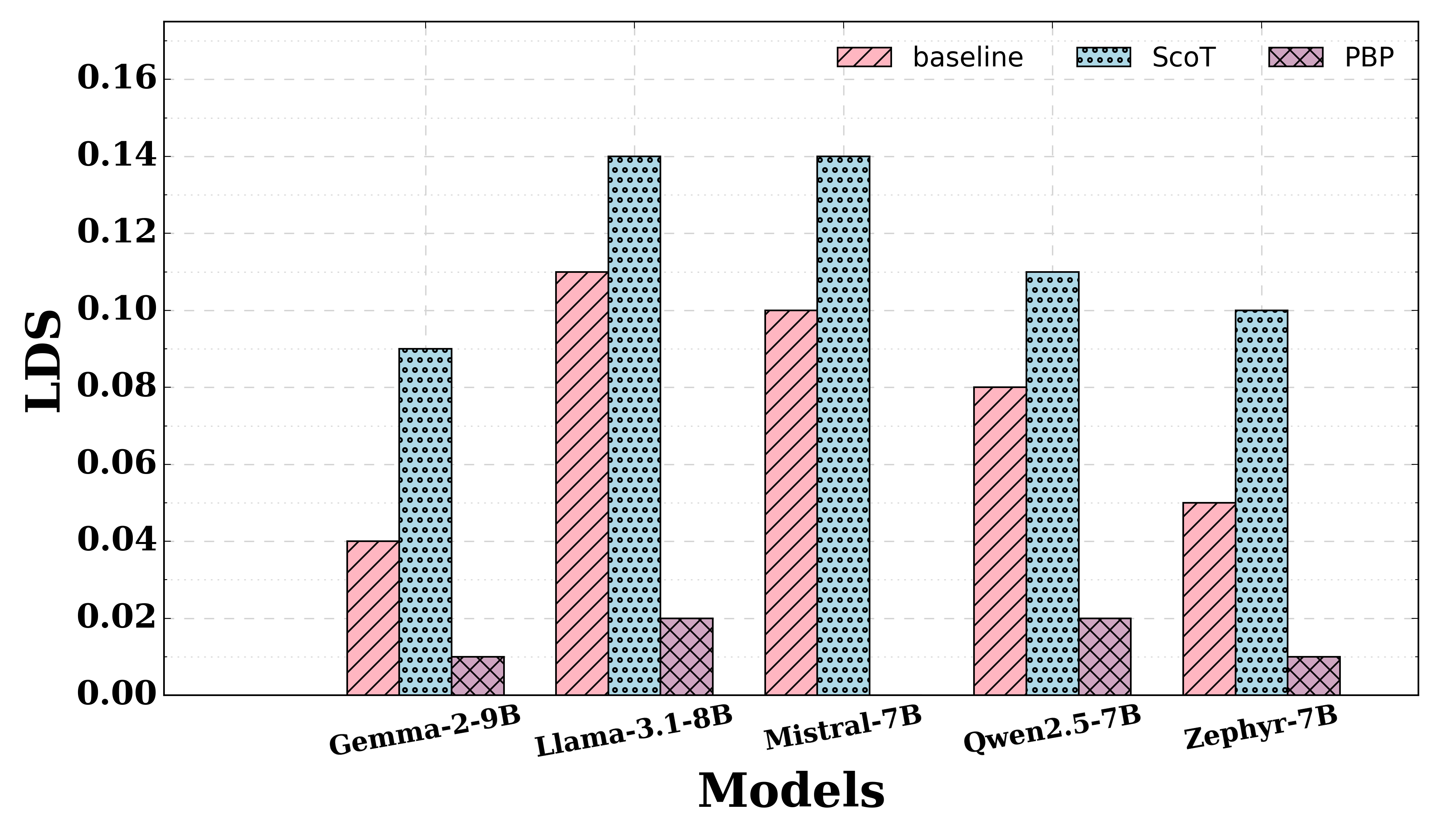}
		\label{f6}	
	}
    \subfigure[Variation in Explanation drift under Same–Decision.]{
		\includegraphics[width=0.22\linewidth]{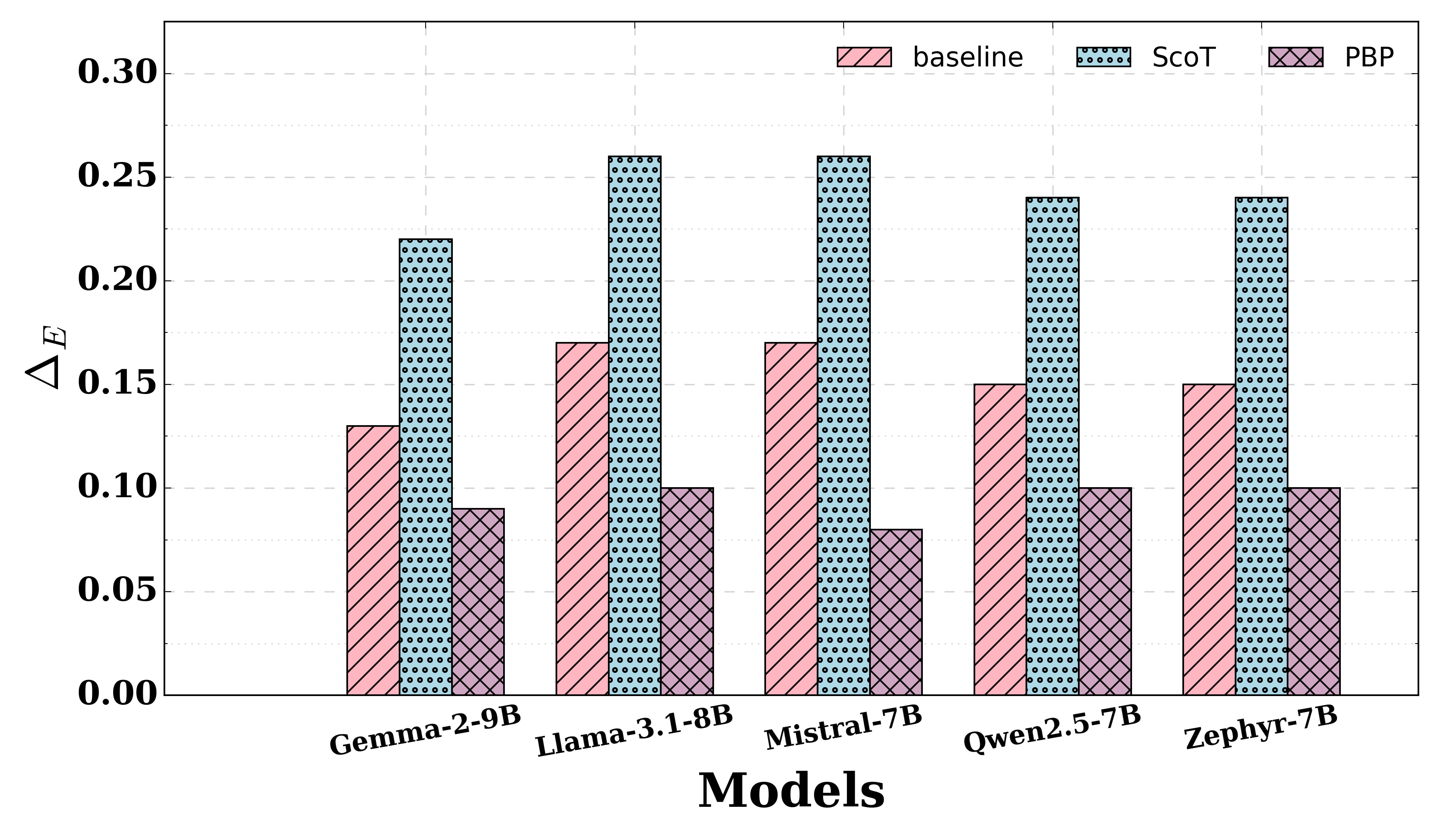}
		\label{f7}	
	}
    \subfigure[Variation in Cue.]{
		\includegraphics[width=0.22\linewidth]{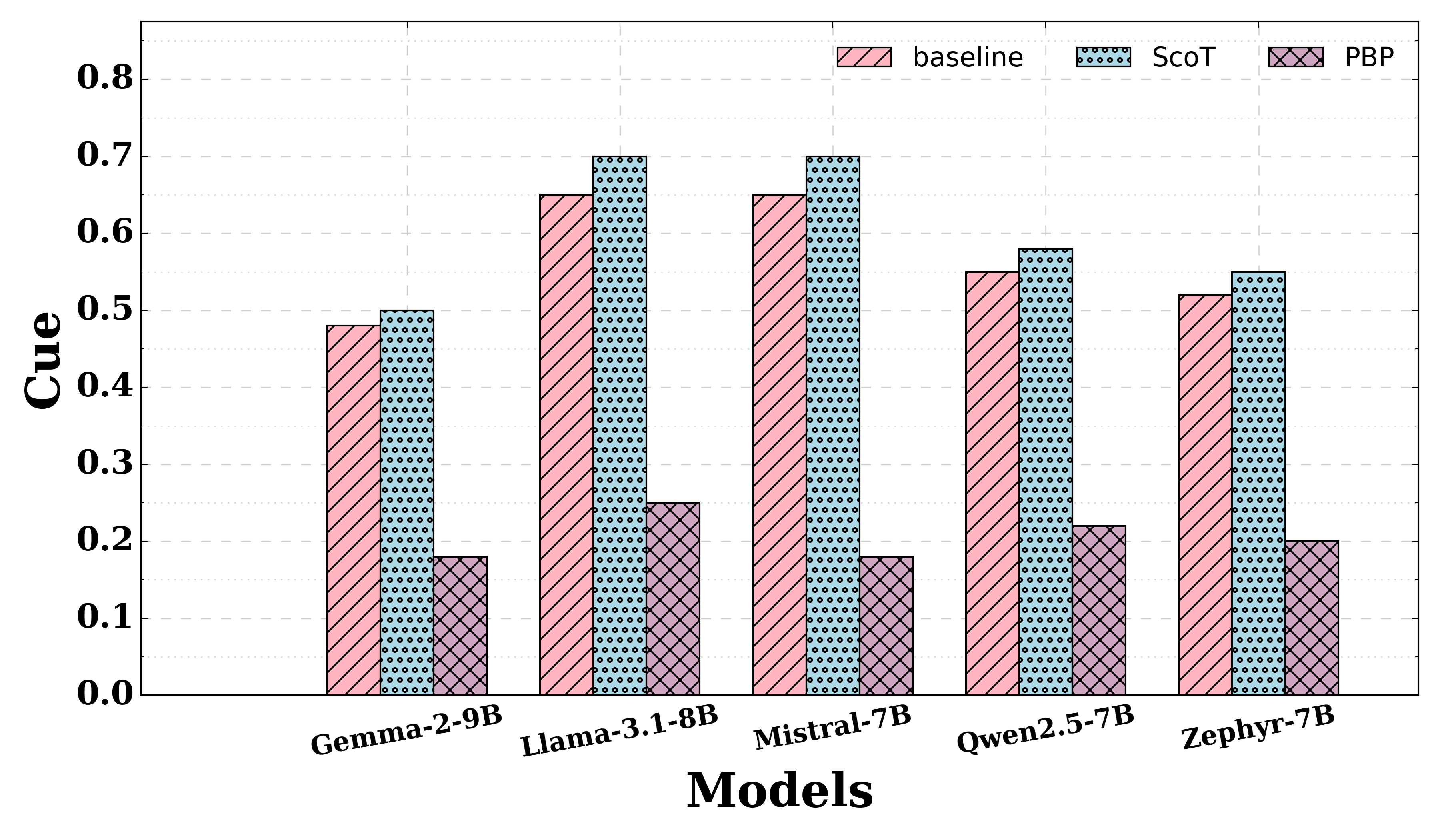}
		\label{f8}	
	}
	\caption{Variation in Different Parameters under Confidence attack.}
	\label{Fig 5}
\end{figure*}
\subsection{Revision Susceptibility after Label Reveal.}
We quantify revision after revealing true labels following an initial judgment under \emph{Flip} labels:
\[
\Pr\!\big[r^{(\mathrm{R})}_{J,d}\neq r^{(\mathrm{F})}_{J,d}\big].
\]
Across models, \textbf{Baseline}/\textbf{SCoT} show high flip-after-reveal rates (75–85\%), while \textbf{PBP} is markedly lower (5–22\%) (refer Fig. \ref{Fig 8}). Concretely, for (Gemma, Llama, Mistral, Qwen, Zephyr): Baseline $\{85,78,76,83,82\}\%$, SCoT $\{80,80,76,82,80\}\%$, PBP $\{22,5,8,18,15\}\%$. This indicates strong label-anchored revision for Baseline/SCoT and robust resistance for PBP.

\subsection{Explanation Rationalization under Same–Decision.}
We evaluate explanation rationalization while holding the verdict fixed using (i) \emph{explanation drift} $\Delta e\!\mid\!\text{Same}$ (refer to Fig. \ref{delta}) (distance between probe and Blind explanations) and (ii) \emph{label–aligned explanation} $\mathrm{LAI}_{\text{text}}\!\mid\!\text{Same}$ (share of label-conforming rhetoric) (refer to Fig. \ref{lai}). Across probes and models, $\Delta e\!\mid\!\text{Same}$ is high ($\approx 0.75$–$0.81$) accorss all models, indicating substantial rewriting even when outcomes do not change. In contrast, $\mathrm{LAI}_{\text{text}}\!\mid\!\text{Same}$ differentiates models: \emph{Mistral-7B} and \emph{Llama-3.1-8B} show the strongest label alignment ($\approx 0.43$ and $0.25$–$0.35$), whereas \emph{Gemma-2-9B}, \emph{Qwen2.5-7B}, and \emph{Zephyr-7B} remain low ($\approx 0.00$–$0.10$). Notably, \textsc{Placebo} yields alignment comparable to \textsc{Flip}/\textsc{True} for the more susceptible models, suggesting increased alignment with presented cues even under Placebo labels. 

\subsection{Robustness under a Verbosity Attack.}
We inflate one candidate with redundant but fluent tokens and evaluate four complementary views of susceptibility: (i) \emph{LAO} (tie–aware, directional outcome anchoring) (refer to Fig. \ref{f1}), (ii) \emph{LDS} (absolute outcome mass moving toward the verbosity–favored side) (refer to Fig. \ref{f2}), (iii) $\Delta e\!\mid\!\text{Same}$ (explanation drift vs.\ Blind on items whose verdict does not change) (refer to Fig. \ref{f3}), and (iv) \emph{Cue–align \%} (share of explanations whose rhetoric aligns with the injected cue) (refer to Fig. \ref{f4}). The four panels collectively show a coherent ordering across models: \textbf{PBP} is most robust, \textbf{SCoT} is most susceptible, and the \textbf{Baseline} lies in between. Concretely, \textbf{PBP} keeps $\emph{LAO}$ low (Gemma 0.15, Llama 0.18, Mistral 0.00, Qwen 0.12, Zephyr 0.10) with \emph{LDS} near zero (0.00–0.02), and its textual changes are modest ($\Delta e\!\mid\!\text{Same}\approx0.08$–$0.11$) with muted cue alignment (15–26\%). The \textbf{Baseline} shows moderate anchoring (e.g., LAO 0.35–0.70; LDS 0.04–0.11), larger rewrites ($\Delta e\!\mid\!\text{Same}\approx0.14$–$0.18$), and higher cue alignment (30–55\%). In contrast, \textbf{SCoT} exhibits pronounced label–directed movement (LAO 0.90–0.97 with LDS 0.09–0.14), the largest narrative drift ($\Delta e\!\mid\!\text{Same}\approx0.22$–$0.28$), and the strongest cue alignment (48–70\%). 

\subsection{Robustness to a Confidence Attack} \label{CA}
We evaluate susceptibility to assertive phrasing by increasing the confidence of one candidate and measuring four indicators: (i) the tie–aware, directional label–anchoring of outcomes $\mathrm{LAO}=\mathrm{LDS}/(\mathrm{LDS}+\mathrm{TS}+\mathrm{OLS})$ (refer to Fig. \ref{f5}); (ii) the absolute label–directed shift in outcome probability $\mathrm{LDS}$ (refer to Fig. \ref{f6}); (iii) the explanation drift under same–decision $\Delta e\!\mid\!\text{Same}$ (distance between probe and Blind explanations, conditional on unchanged verdicts) (refer to Fig. \ref{f7}); and (iv) the fraction of cue–aligned explanations (\emph{Cue}) (refer to Fig. \ref{f8}). Across models, the ordering is consistent and statistically salient: \textbf{PBP} exhibits the lowest outcome anchoring and minimal text drift, the \textbf{Baseline} shows moderate effects, and \textbf{SCoT} is most affected. Concretely, under PBP, $\mathrm{LAO}$ remains small (approximately $0.00$–$0.15$) with near–zero $\mathrm{LDS}$ ($0.00$–$0.02$), while explanations change only modestly ($\Delta e\!\approx\!0.08$–$0.11$) and cue alignment is limited ($\approx\!18$–$26\%$). The Baseline exhibits intermediate susceptibility ($\mathrm{LAO}\!\approx\!0.35$–$0.65$, $\mathrm{LDS}\!\approx\!0.04$–$0.11$, $\Delta e\!\approx\!0.13$–$0.18$, Cue $\approx\!48$–$67\%$). By contrast, SCoT displays pronounced anchoring and rationalization ($\mathrm{LAO}\!\approx\!0.90$–$0.96$, $\mathrm{LDS}\!\approx\!0.08$–$0.14$, $\Delta e\!\approx\!0.22$–$0.28$, Cue $\approx\!50$–$70\%$), indicating that rubric dimensions such as fluency/authority are over–rewarded by assertive tone.


\section{Limitations and Scope}

This paper targets a specific reliability failure in LLM-based evaluation, \emph{cue-driven rationalization}, where decisions and explanations change under non-evidential cues while the underlying texts remain fixed. We frame \emph{cue invariance} as a necessary (but not sufficient) condition for reliable explanations and use controlled interventions to diagnose violations of this condition, rather than to fully validate explanation faithfulness. Our explanation-level metrics function as relative diagnostics under fixed evidence and are not intended as ground-truth semantic judgments; their behavior may depend on the choice of scorer. Empirically, we focus on pairwise summarization and report aggregate effects without confidence intervals, multi-seed variance, or measured cost/latency; the \textsc{PBP} protocol also increases abstention when candidates are near-equivalent and introduces modest, predictable overhead due to serialized prompting. These constraints reflect deliberate design trade-offs to isolate a clean causal signal, and motivate future work on cross-task generalization, uncertainty estimation, human validation of explanation alignment, and cost–robustness analysis.

\section{Conclusion}

In this paper, we investigated whether explanations provided by LLM judges were faithful to their decision-making or served as post-hoc rationalizations. We introduced a causal probing framework that manipulated metadata labels and allowed us to isolate the influence of external cues on explanations. We defined a suite of metrics to measure fabrication, label alignment, cue susceptibility, stereotype intrusion, and consistency. We further designed rationalization attacks that exposed systematic vulnerabilities and proposed two mitigation strategies: structured chain-of-thought prompting and PBP. Our experiments showed that LLM judges frequently rationalized under label perturbations, but that our mitigation strategies reduced rationalization bias significantly. Overall, this work established a foundation for auditing and mitigating explanation faithfulness in LLM judges.

However, our evaluation is limited in scope and statistical depth: we study summarization only, rely on explanation-level metrics that depend on specific rationale scorers, and do not report confidence intervals or multi-seed variance. These are principled deferrals; future work will test transfer to tasks with tighter contextual control, compare rationale scorers to disentangle semantic and stylistic effects, and measure cost-robustness trade-offs across models and serving stacks.

\bibliographystyle{IEEEtran}
\bibliography{ref}

@inproceedings{10.5555/3666122.3668142,
author = {Zheng, Lianmin and Chiang, Wei-Lin and Sheng, Ying and Zhuang, Siyuan and Wu, Zhanghao and Zhuang, Yonghao and Lin, Zi and Li, Zhuohan and Li, Dacheng and Xing, Eric P. and Zhang, Hao and Gonzalez, Joseph E. and Stoica, Ion},
title = {Judging LLM-as-a-judge with MT-bench and Chatbot Arena},
year = {2023},
publisher = {Curran Associates Inc.},
address = {Red Hook, NY, USA},
booktitle = {Proceedings of the 37th International Conference on Neural Information Processing Systems},
articleno = {2020},
numpages = {29},
location = {New Orleans, LA, USA},
series = {NIPS '23},
url={https://neurips.cc/virtual/2023/poster/73434}
}

@inproceedings{wu-aji-2025-style,
    title = "Style Over Substance: Evaluation Biases for Large Language Models",
    author = "Wu, Minghao  and
      Aji, Alham Fikri",
    editor = "Rambow, Owen  and
      Wanner, Leo  and
      Apidianaki, Marianna  and
      Al-Khalifa, Hend  and
      Eugenio, Barbara Di  and
      Schockaert, Steven",
    booktitle = "Proceedings of the 31st International Conference on Computational Linguistics",
    month = jan,
    year = "2025",
    address = "Abu Dhabi, UAE",
    publisher = "Association for Computational Linguistics",
    url = "https://aclanthology.org/2025.coling-main.21/",
    pages = "297--312",
}

@inproceedings{koo-etal-2024-benchmarking,
    title = "Benchmarking Cognitive Biases in Large Language Models as Evaluators",
    author = "Koo, Ryan  and
      Lee, Minhwa  and
      Raheja, Vipul  and
      Park, Jong Inn  and
      Kim, Zae Myung  and
      Kang, Dongyeop",
    editor = "Ku, Lun-Wei  and
      Martins, Andre  and
      Srikumar, Vivek",
    booktitle = "Findings of the Association for Computational Linguistics: ACL 2024",
    month = aug,
    year = "2024",
    address = "Bangkok, Thailand",
    publisher = "Association for Computational Linguistics",
    url = "https://aclanthology.org/2024.findings-acl.29/",
    doi = "10.18653/v1/2024.findings-acl.29",
    pages = "517--545",
}

@inproceedings{ye2025justice,
title={Justice or Prejudice? Quantifying Biases in {LLM}-as-a-Judge},
author={Jiayi Ye and Yanbo Wang and Yue Huang and Dongping Chen and Qihui Zhang and Nuno Moniz and Tian Gao and Werner Geyer and Chao Huang and Pin-Yu Chen and Nitesh V Chawla and Xiangliang Zhang},
booktitle={The Thirteenth International Conference on Learning Representations},
year={2025},
url={https://openreview.net/forum?id=3GTtZFiajM}
}

@inproceedings{hgg,
author = {Lee, Dongryeol and Hwang, Yerin and Kim, Yongil and Park, Joonsuk and Jung, Kyomin},
year = {2025},
month = {01},
pages = {8962-8984},
title = {Are LLM-Judges Robust to Expressions of Uncertainty? Investigating the effect of Epistemic Markers on LLM-based Evaluation},
doi = {10.18653/v1/2025.naacl-long.452},
publisher={Proceedings of the 2025 Conference of the Nations of the Americas Chapter of the Association for Computational Linguistics: Human Language Technologies},
pages={8962–8984},
url={https://aclanthology.org/2025.naacl-long.452.pdf}
}

@inproceedings{raina-etal-2024-llm,
    title = "Is {LLM}-as-a-Judge Robust? Investigating Universal Adversarial Attacks on Zero-shot {LLM} Assessment",
    author = "Raina, Vyas  and
      Liusie, Adian  and
      Gales, Mark",
    editor = "Al-Onaizan, Yaser  and
      Bansal, Mohit  and
      Chen, Yun-Nung",
    booktitle = "Proceedings of the 2024 Conference on Empirical Methods in Natural Language Processing",
    month = nov,
    year = "2024",
    address = "Miami, Florida, USA",
    publisher = "Association for Computational Linguistics",
    url = "https://aclanthology.org/2024.emnlp-main.427/",
    doi = "10.18653/v1/2024.emnlp-main.427",
    pages = "7499--7517",

}

@inproceedings{chen-etal-2024-humans,
    title = "Humans or {LLM}s as the Judge? A Study on Judgement Bias",
    author = "Chen, Guiming Hardy  and
      Chen, Shunian  and
      Liu, Ziche  and
      Jiang, Feng  and
      Wang, Benyou",
    editor = "Al-Onaizan, Yaser  and
      Bansal, Mohit  and
      Chen, Yun-Nung",
    booktitle = "Proceedings of the 2024 Conference on Empirical Methods in Natural Language Processing",
    month = nov,
    year = "2024",
    address = "Miami, Florida, USA",
    publisher = "Association for Computational Linguistics",
    url = "https://aclanthology.org/2024.emnlp-main.474/",
    doi = "10.18653/v1/2024.emnlp-main.474",
    pages = "8301--8327",
}

@inproceedings{li-etal-2024-split,
    title = "Split and Merge: Aligning Position Biases in {LLM}-based Evaluators",
    author = "Li, Zongjie  and
      Wang, Chaozheng  and
      Ma, Pingchuan  and
      Wu, Daoyuan  and
      Wang, Shuai  and
      Gao, Cuiyun  and
      Liu, Yang",
    editor = "Al-Onaizan, Yaser  and
      Bansal, Mohit  and
      Chen, Yun-Nung",
    booktitle = "Proceedings of the 2024 Conference on Empirical Methods in Natural Language Processing",
    month = nov,
    year = "2024",
    address = "Miami, Florida, USA",
    publisher = "Association for Computational Linguistics",
    url = "https://aclanthology.org/2024.emnlp-main.621/",
    doi = "10.18653/v1/2024.emnlp-main.621",
    pages = "11084--11108",

}

@inproceedings{thakur-etal-2025-judging,
    title = "Judging the Judges: Evaluating Alignment and Vulnerabilities in {LLM}s-as-Judges",
    author = "Thakur, Aman Singh  and
      Choudhary, Kartik  and
      Ramayapally, Venkat Srinik  and
      Vaidyanathan, Sankaran  and
      Hupkes, Dieuwke",

    booktitle = "Proceedings of the Fourth Workshop on Generation, Evaluation and Metrics (GEM{\texttwosuperior})",
    month = jul,
    year = "2025",
    address = "Vienna, Austria and virtual meeting",
    publisher = "Association for Computational Linguistics",
    url = "https://aclanthology.org/2025.gem-1.33/",
    pages = "404--430",
    ISBN = "979-8-89176-261-9",

}

@inproceedings{10.1145/3708359.3712091,
author = {Szymanski, Annalisa and Ziems, Noah and Eicher-Miller, Heather A. and Li, Toby Jia-Jun and Jiang, Meng and Metoyer, Ronald A.},
title = {Limitations of the LLM-as-a-Judge Approach for Evaluating LLM Outputs in Expert Knowledge Tasks},
year = {2025},
isbn = {9798400713064},
publisher = {Association for Computing Machinery},
address = {New York, NY, USA},
url = {https://doi.org/10.1145/3708359.3712091},
doi = {10.1145/3708359.3712091},
booktitle = {Proceedings of the 30th International Conference on Intelligent User Interfaces},
pages = {952–966},
numpages = {15},
location = {
},
series = {IUI '25}
}

@article{Croxford2025LLMEval,
  author    = {Croxford, E. and Gao, Y. and Pellegrino, N. and others},
  title     = {Current and future state of evaluation of large language models for medical summarization tasks},
  journal   = {npj Health Systems},
  volume    = {2},
  number    = {6},
  pages     = {6},
  year      = {2025},
  month     = {Feb},
  doi       = {10.1038/s44401-024-00011-2},
  url       = {https://doi.org/10.1038/s44401-024-00011-2}
}

@inproceedings{Turpin2023UnfaithfulCoT,
  author    = {Miles Turpin and Julian Michael and Ethan Perez and Samuel R. Bowman},
  title     = {Language Models Don't Always Say What They Think: Unfaithful Explanations in Chain-of-Thought Prompting},
  booktitle = {Advances in Neural Information Processing Systems (NIPS 2023), Poster},
  year      = {2023},
  note      = {Poster},
  url={https://dl.acm.org/doi/10.5555/3666122.3669397}
}

@article{Chen2025ReasoningModelsDontAlwaysSay,
  title   = {Reasoning Models Don’t Always Say What They Think},
  author  = {Yanda Chen and Joe Benton and Ansh Radhakrishnan and Jonathan Uesato and Carson Denison and John Schulman and Arushi Somani and Peter Hase and Misha Wagner and Fabien Roger and Vlad Mikulik and Sam Bowman and Jan Leike and Jared Kaplan and Ethan Perez and Alignment Science Team, Anthropic},
  year    = {2025},
  journal = {Anthropic Research Report},
  note    = {Working paper; available at Anthropic's website under “Reasoning Models Don’t Always Say What They Think”},
  url     = {https://assets.anthropic.com/m/71876fabef0f0ed4/original/reasoning_models_paper.pdf}
}

@misc{lanham2023measuringfaithfulnesschainofthoughtreasoning,
      title={Measuring Faithfulness in Chain-of-Thought Reasoning}, 
      author={Tamera Lanham and Anna Chen and Ansh Radhakrishnan and Benoit Steiner and Carson Denison and Danny Hernandez and Dustin Li and Esin Durmus and Evan Hubinger and Jackson Kernion and Kamilė Lukošiūtė and Karina Nguyen and Newton Cheng and Nicholas Joseph and Nicholas Schiefer and Oliver Rausch and Robin Larson and Sam McCandlish and Sandipan Kundu and Saurav Kadavath and Shannon Yang and Thomas Henighan and Timothy Maxwell and Timothy Telleen-Lawton and Tristan Hume and Zac Hatfield-Dodds and Jared Kaplan and Jan Brauner and Samuel R. Bowman and Ethan Perez},
      year={2023},
      eprint={2307.13702},
      archivePrefix={arXiv},
      primaryClass={cs.AI},
      url={https://arxiv.org/abs/2307.13702}, 
}

@misc{lewislim2025analysingchainthoughtdynamics,
      title={Analysing Chain of Thought Dynamics: Active Guidance or Unfaithful Post-hoc Rationalisation?}, 
      author={Samuel Lewis-Lim and Xingwei Tan and Zhixue Zhao and Nikolaos Aletras},
      year={2025},
      eprint={2508.19827},
      archivePrefix={arXiv},
      primaryClass={cs.AI},
      url={https://arxiv.org/abs/2508.19827}, 
}

@misc{chuang2024faithlmfaithfulexplanationslarge,
      title={FaithLM: Towards Faithful Explanations for Large Language Models}, 
      author={Yu-Neng Chuang and Guanchu Wang and Chia-Yuan Chang and Ruixiang Tang and Shaochen Zhong and Fan Yang and Mengnan Du and Xuanting Cai and Xia Hu},
      year={2024},
      eprint={2402.04678},
      archivePrefix={arXiv},
      primaryClass={cs.CL},
      url={https://arxiv.org/abs/2402.04678}, 
}

@inproceedings{li-etal-2025-drift,
    title = "Drift: Enhancing {LLM} Faithfulness in Rationale Generation via Dual-Reward Probabilistic Inference",
    author = "Li, Jiazheng  and
      Yan, Hanqi  and
      He, Yulan",
    editor = "Che, Wanxiang  and
      Nabende, Joyce  and
      Shutova, Ekaterina  and
      Pilehvar, Mohammad Taher",
    booktitle = "Proceedings of the 63rd Annual Meeting of the Association for Computational Linguistics (Volume 1: Long Papers)",
    month = jul,
    year = "2025",
    address = "Vienna, Austria",
    publisher = "Association for Computational Linguistics",
    url = "https://aclanthology.org/2025.acl-long.340/",
    doi = "10.18653/v1/2025.acl-long.340",
    pages = "6850--6866",
    ISBN = "979-8-89176-251-0",
}

@article{INR-015,
url = {http://dx.doi.org/10.1561/1500000015},
year = {2011},
volume = {5},
journal = {Foundations and Trends® in Information Retrieval},
title = {Automatic Summarization},
doi = {10.1561/1500000015},
issn = {1554-0669},
number = {2–3},
pages = {103-233},
author = {Ani Nenkova and Kathleen McKeown}
}

@inproceedings{Desmond2025EvalAssist,
  author    = {Michael Desmond and Zahra Ashktorab and Werner Geyer and Elizabeth M. Daly and Mart{\'i}n Santill{\'a}n Cooper and Qian Pan and Rahul Nair and Nico Wagner and Tejaswini Pedapati},
  title     = {EvalAssist: LLM-as-a-Judge Simplified},
  booktitle = {Proceedings of the AAAI Conference on Artificial Intelligence, Demonstration Track},
  volume    = {39},
  number    = {28},
  pages     = {35351},
  year      = {2025},
  publisher = {AAAI Press},
  doi       = {10.1609/aaai.v39i28.35351},
  url       = {https://doi.org/10.1609/aaai.v39i28.35351}
}

@article{10.1162/tacl_a_00373,
    author = {Fabbri, Alexander R. and Kryściński, Wojciech and McCann, Bryan and Xiong, Caiming and Socher, Richard and Radev, Dragomir},
    title = {SummEval: Re-evaluating Summarization Evaluation},
    journal = {Transactions of the Association for Computational Linguistics},
    volume = {9},
    pages = {391-409},
    year = {2021},
    month = {04},
    issn = {2307-387X},
    doi = {10.1162/tacl_a_00373},
    url = {https://doi.org/10.1162/tacl_a_00373},
    eprint = {https://direct.mit.edu/tacl/article-pdf/doi/10.1162/tacl_a_00373/1923949/tacl_a_00373.pdf},
}

@inproceedings{10.5555/3524938.3525989,
author = {Zhang, Jingqing and Zhao, Yao and Saleh, Mohammad and Liu, Peter J.},
title = {PEGASUS: pre-training with extracted gap-sentences for abstractive summarization},
year = {2020},
publisher = {JMLR.org},
booktitle = {Proceedings of the 37th International Conference on Machine Learning},
articleno = {1051},
numpages = {12},
series = {ICML'20},
url={https://dl.acm.org/doi/abs/10.5555/3524938.3525989}
}

@inproceedings{factual_1,
title    = {On Faithfulness and Factuality in Abstractive Summarization},
author    = {Joshua Maynez and Shashi Narayan and Bernd Bohnet and Ryan Thomas Mcdonald},
year    = {2020},
booktitle    = {Proceedings of The 58th Annual Meeting of the Association for Computational Linguistics (ACL)},
url={https://aclanthology.org/2020.acl-main.173.pdf}}

@inproceedings{factual_2,
    title = "Annotating and Modeling Fine-grained Factuality in Summarization",
    author = "Goyal, Tanya  and
      Durrett, Greg",
    editor = "Toutanova, Kristina  and
      Rumshisky, Anna  and
      Zettlemoyer, Luke  and
      Hakkani-Tur, Dilek  and
      Beltagy, Iz  and
      Bethard, Steven  and
      Cotterell, Ryan  and
      Chakraborty, Tanmoy  and
      Zhou, Yichao",
    booktitle = "Proceedings of the 2021 Conference of the North American Chapter of the Association for Computational Linguistics: Human Language Technologies",
    month = jun,
    year = "2021",
    address = "Online",
    publisher = "Association for Computational Linguistics",
    url = "https://aclanthology.org/2021.naacl-main.114/",
    doi = "10.18653/v1/2021.naacl-main.114",
    pages = "1449--1462"
}

@inproceedings{completeness_1,
    title = "Evaluating Content Selection in Summarization: The Pyramid Method",
    author = "Nenkova, Ani  and
      Passonneau, Rebecca",
    booktitle = "Proceedings of the Human Language Technology Conference of the North {A}merican Chapter of the Association for Computational Linguistics: {HLT}-{NAACL} 2004",
    month = may # " 2 - " # may # " 7",
    year = "2004",
    address = "Boston, Massachusetts, USA",
    publisher = "Association for Computational Linguistics",
    url = "https://aclanthology.org/N04-1019/",
    pages = "145--152"
}

@inproceedings{completness_2,
    title = "Automatic Evaluation of Summaries Using N-gram Co-occurrence Statistics",
    author = "Lin, Chin-Yew  and
      Hovy, Eduard",
    booktitle = "Proceedings of the 2003 Human Language Technology Conference of the North {A}merican Chapter of the Association for Computational Linguistics",
    year = "2003",
    url = "https://aclanthology.org/N03-1020/",
    pages = "150--157"
}

@article{coherence_2,
  title={Ffci: A framework for interpretable automatic evaluation of summarization},
  author={Koto, Fajri and Baldwin, Timothy and Lau, Jey Han},
  journal={Journal of Artificial Intelligence Research},
  volume={73},
  pages={1553--1607},
  year={2022},
  url={https://dl.acm.org/doi/10.1613/jair.1.13167}
}

@inproceedings{hermann2015teaching,
  title={Teaching Machines to Read and Comprehend},
  author={Hermann, Karl Moritz and Kocisky, Tomas and Grefenstette, Edward and Espeholt, Lasse and Kay, Will and Suleyman, Mustafa and Blunsom, Phil},
  booktitle={Advances in Neural Information Processing Systems (NeurIPS)},
  volume={28},
  year={2015}
}

@inproceedings{narayan2018don,
  title={Don't Give Me the Details, Just the Summary! Topic-Aware Convolutional Neural Networks for Extreme Summarization},
  author={Narayan, Shashi and Cohen, Shay B. and Lapata, Mirella},
  booktitle={Proceedings of the 2018 Conference on Empirical Methods in Natural Language Processing (EMNLP)},
  pages={1797--1807},
  year={2018}
}

\appendix

\section*{Category \& Likert Scale}

\begin{table*}[htbp]
\centering
\caption{Summary evaluation categories and their definitions.}
\begin{tabular}{p{3.5cm} p{9.5cm}}
\hline
\textbf{Category} & \textbf{Definition} \\ \hline

\textbf{Factual Accuracy} & 
Measures how faithfully the summary reflects the information in the source document, without introducing hallucinations, distortions, or fabricated details. \\[0.3cm]

\textbf{Completeness (Content Coverage)} & 
Evaluates how well the summary captures the key information, main points, and essential content units of the source document. \\[0.3cm]

\textbf{Coherence \& Fluency} & 
Assesses the linguistic quality of the summary, including clarity, grammatical correctness, logical flow, and readability. \\

\hline
\end{tabular}
\label{tab:summary_categories}
\end{table*}

\paragraph{Rationale for Choosing Evaluation Categories.}
We evaluate summaries using three core dimensions: Factual Accuracy, 
Completeness, and Coherence \& Fluency. 
Factual Accuracy is essential because modern abstractive models often generate 
fluent but incorrect statements; prior work shows that factual consistency is a crucial 
determinant of trustworthy summarization \cite{factual_1,factual_2}. 
Completeness ensures that summaries retain key content elements from the source, 
aligned with classical frameworks such as the Pyramid Method and ROUGE, which emphasise 
coverage of important information \cite{completeness_1,completness_2}. 
Coherence \& Fluency capture the readability and structural quality of the 
summary, reflecting grammaticality, clarity, and logical flow, dimensions shown to 
correlate strongly with human preferences in recent evaluation benchmarks 
\cite{coherence_2}. 
Together, these categories form a comprehensive and balanced framework for assessing 
both content and linguistic quality in summarization.

\paragraph{Scoring Rubric:}
For each evaluation category, we adopt a Likert-scale scoring rubric ranging 
from 1 to 5, where 1 denotes the lowest quality and 5 denotes the highest. 
This scale provides fine-grained resolution while remaining intuitive for both human 
evaluators and LLM-based judges. A score of 1 indicates severe deficiencies, such as 
major factual errors, missing essential content, or incoherent writing, whereas a score 
of 5 reflects excellent factual grounding, comprehensive content coverage, and highly 
fluent and well-structured summaries. The 1--5 scale is widely used in summarization 
evaluation because it offers a reliable and interpretable method for comparing models 
across multiple dimensions.

\section{LLM Judge Prompt.}
We instruct the LLM to evaluate a candidate summary with respect to three dimensions:
Factual Accuracy, Completeness, and Coherence \& Fluency.
The model is explicitly informed of the definition of each dimension and is required to
assign a score from 1 to 5 (1 = worst, 5 = best) using a Likert-scale rubric. The full
prompt is shown below.

\begin{quote}
You are an expert evaluator of text summarization quality. You will be given:

1. A \textbf{Source Document}, and  
2. A \textbf{Candidate Summary}.  

Your task is to evaluate the quality of the summary using the three criteria defined
below. Assign a score from 1 to 5 for each criterion, where 1 indicates very poor 
performance and 5 indicates excellent performance.

\textbf{Evaluation Criteria}

\begin{itemize}
    \item \textbf{Factual Accuracy}: 
    Assess how faithfully the summary reflects the information in the source document. 
    The summary should not introduce hallucinated content, distort facts, or omit
    critical causal or factual relationships.

    \item \textbf{Completeness (Content Coverage)}:
    Evaluate how well the summary captures the key information, major points, and 
    essential content units from the source document. Consider whether important 
    content is missing.

    \item \textbf{Coherence \& Fluency}:
    Assess the linguistic quality of the summary. The summary should be grammatically 
    correct, logically structured, clear, and easy to read. Sentences should flow 
    smoothly and maintain a consistent style.

\end{itemize}

\textbf{Scoring Rubric (1--5 Scale)}

\begin{itemize}
    \item \textbf{1}: Very poor. Major errors, severe factual inconsistencies, missing 
    key content, or extremely unclear writing.
    \item \textbf{2}: Poor. Multiple issues, limited content coverage, or low fluency.
    \item \textbf{3}: Acceptable. Partially correct and somewhat informative, but with 
    noticeable inaccuracies, omissions, or awkward writing.
    \item \textbf{4}: Good. Mostly correct, captures most key information, and reads 
    well with minor issues.
    \item \textbf{5}: Excellent. Fully accurate, complete, and highly fluent with no 
    notable errors.
\end{itemize}

\section*{Ablation study}
\subsection{Human Check for Near-Equivalent Pairs}
To confirm that many instances are genuinely hard to separate in overall quality, we conducted a small human check. We randomly sampled $n = 100$ documents for manual review. 

Two annotators independently compared the two summaries for each sampled document and gave a single overall label: either selecting the better summary or marking \emph{Tie} when neither summary was clearly better. Annotator~1 marked \emph{Tie} on $94$ of the $100$ items, and Annotator~2 marked \emph{Tie} on $92$ of the $100$ items. These results support our assumption that a substantial portion of our comparisons are near-equivalent, which motivates tie-aware analysis.

\subsection{Blind Judgments (no labels revealed).}
Table~\ref{tab:blind-counts} reports raw Blind counts per model and scheme for \emph{No Selection} (Tie) and strict pairings $[1,2]$ and $[2,1]$. The pattern is consistent across judges: \textbf{PBP} yields the largest abstention rates (e.g., Gemma 937/1000; Llama 924/1000; Mistral 824/1000; Qwen 885/1000; Zephyr 897/1000), \textbf{SCoT} attains substantial but lower tie rates (Gemma 840/1000; Llama 804/1000; Mistral 710/1000; Qwen 728/1000; Zephyr 735/1000), and the \textbf{Baseline} never abstains (all zeros in the \emph{No Selection} column). Consequently, equality detection $\mathrm{EDR}_B$ is highest for PBP (approx 0.82–0.94), moderate for SCoT (approx 0.71–0.84), and zero for Baseline. The residual non–tie choices under SCoT/PBP are roughly balanced (e.g., Gemma SCoT: $97$ vs.\ $63$; PBP: $34$ vs.\ $29$), leading to very small neutrality deviation $\mathrm{ND}_B$ for PBP (approx 0.001–0.005) and modest values for SCoT (approx 0.019–0.122), while Baseline shows skew because it is forced to pick a side (e.g., Gemma $540$ vs.\ $460$ over 1000 items). Overall, the Blind counts already separate the schemes: PBP exhibits the highest abstention rates under Blind conditions, SCoT helps but still finds small differences, and Baseline cannot abstain.

\begin{table*}[!htbp]
\centering
\caption{Blind (no-label) decisions by scheme and model. Each cell reports raw counts of \emph{No Selection} (Tie), $[1,2]$ (LLM $\succ$ TradML), and $[2,1]$ (TradML $\succ$ LLM). Totals per model are $1000$ for SCoT and PBP and the Baseline. The consistent ordering, PBP $\gg$ SCoT $\gg$ Baseline in \emph{No Selection}, implies highest equality detection and lowest neutrality deviation for PBP.}
\label{tab:blind-counts}

\begin{tabular}{|l|lll|lll|lll|}
\hline
\multirow{2}{*}{Model (Judge)} & \multicolumn{3}{l|}{Baseline}                                          & \multicolumn{3}{l|}{SCoT}                                              & \multicolumn{3}{l|}{PBP}                                               \\ \cline{2-10} 
                               & \multicolumn{1}{l|}{No Selection} & \multicolumn{1}{l|}{[1,2]} & [2,1] & \multicolumn{1}{l|}{No Selection} & \multicolumn{1}{l|}{[1,2]} & [2,1] & \multicolumn{1}{l|}{No Selection} & \multicolumn{1}{l|}{[1,2]} & [2,1] \\ \hline
Gemma-2-9B                     & \multicolumn{1}{l|}{0}            & \multicolumn{1}{l|}{540}   & 460   & \multicolumn{1}{l|}{840}          & \multicolumn{1}{l|}{97}    & 63    & \multicolumn{1}{l|}{937}          & \multicolumn{1}{l|}{34}    & 29    \\ \hline
Llama-3.1-8B                   & \multicolumn{1}{l|}{0}            & \multicolumn{1}{l|}{470}   & 530   & \multicolumn{1}{l|}{804}          & \multicolumn{1}{l|}{110}   & 86    & \multicolumn{1}{l|}{924}          & \multicolumn{1}{l|}{37}    & 39    \\ \hline
Mistral-7B                     & \multicolumn{1}{l|}{0}            & \multicolumn{1}{l|}{550}   & 450   & \multicolumn{1}{l|}{710}          & \multicolumn{1}{l|}{206}   & 84    & \multicolumn{1}{l|}{824}          & \multicolumn{1}{l|}{93}    & 88    \\ \hline
Qwen2.5-7B                     & \multicolumn{1}{l|}{0}            & \multicolumn{1}{l|}{460}   & 540   & \multicolumn{1}{l|}{728}          & \multicolumn{1}{l|}{168}   & 104   & \multicolumn{1}{l|}{885}          & \multicolumn{1}{l|}{58}    & 57    \\ \hline
Zephyr-7B                      & \multicolumn{1}{l|}{0}            & \multicolumn{1}{l|}{510}   & 490   & \multicolumn{1}{l|}{735}          & \multicolumn{1}{l|}{142}   & 123   & \multicolumn{1}{l|}{897}          & \multicolumn{1}{l|}{50}    & 53    \\ \hline
\end{tabular}
\end{table*}

\begin{table*}[!htbp]
\centering
\caption{True-label decisions by scheme and model. Counts for \emph{No Selection} (Tie) and strict pairings $[1,2]$ (LLM $\succ$ TradML) and $[2,1]$ (TradML $\succ$ LLM). Baseline exhibits maximal label anchoring (no ties, heavy $[1,2]$), SCoT shows reduced but still substantial anchoring with limited abstention, and PBP sustains high abstention with near-balanced non–tie choices, reflecting minimal outcome-level label susceptibility.}
\label{tab:true-counts}

\begin{tabular}{|l|lll|lll|lll|}
\hline
\multirow{2}{*}{Model (Judge)} & \multicolumn{3}{l|}{Baseline}                                          & \multicolumn{3}{l|}{SCoT}                                              & \multicolumn{3}{l|}{PBP}                                               \\ \cline{2-10} 
                               & \multicolumn{1}{l|}{No Selection} & \multicolumn{1}{l|}{[1,2]} & [2,1] & \multicolumn{1}{l|}{No Selection} & \multicolumn{1}{l|}{[1,2]} & [2,1] & \multicolumn{1}{l|}{No Selection} & \multicolumn{1}{l|}{[1,2]} & [2,1] \\ \hline
Gemma-2-9B                     & \multicolumn{1}{l|}{0}            & \multicolumn{1}{l|}{1000}  & 0     & \multicolumn{1}{l|}{215}          & \multicolumn{1}{l|}{680}   & 105   & \multicolumn{1}{l|}{925}          & \multicolumn{1}{l|}{39}    & 36    \\ \hline
Llama-3.1-8B                   & \multicolumn{1}{l|}{0}            & \multicolumn{1}{l|}{860}   & 140   & \multicolumn{1}{l|}{198}          & \multicolumn{1}{l|}{660}   & 142   & \multicolumn{1}{l|}{920}          & \multicolumn{1}{l|}{42}    & 38    \\ \hline
Mistral-7B                     & \multicolumn{1}{l|}{0}            & \multicolumn{1}{l|}{1000}  & 0     & \multicolumn{1}{l|}{76}           & \multicolumn{1}{l|}{920}   & 4     & \multicolumn{1}{l|}{825}          & \multicolumn{1}{l|}{90}    & 85    \\ \hline
Qwen2.5-7B                     & \multicolumn{1}{l|}{0}            & \multicolumn{1}{l|}{750}   & 250   & \multicolumn{1}{l|}{103}          & \multicolumn{1}{l|}{867}   & 30    & \multicolumn{1}{l|}{872}          & \multicolumn{1}{l|}{68}    & 60    \\ \hline
Zephyr-7B                      & \multicolumn{1}{l|}{0}            & \multicolumn{1}{l|}{1000}  & 0     & \multicolumn{1}{l|}{128}          & \multicolumn{1}{l|}{805}   & 67    & \multicolumn{1}{l|}{895}          & \multicolumn{1}{l|}{51}    & 54    \\ \hline
\end{tabular}
\end{table*}

\subsection{Decisions with True Labels Revealed.}
Table~\ref{tab:true-counts} reports raw counts when the correct labels are shown. The \textbf{Baseline} is maximally label–anchored: it never abstains and selects $[1,2]$ almost exclusively (e.g., Gemma/Mistral/Zephyr $1000\!:\!0$, Llama $860\!:\!140$, Qwen $750\!:\!250$). \textbf{SCoT} introduces some abstention (No Selection $76$–$215$) but still strongly favors the label–advantaged option (e.g., Gemma $680\!:\!105$, Mistral $920\!:\!4$), yielding large label–directed shifts. In contrast, \textbf{PBP} maintains high abstention (No Selection $825$–$925$) with small, nearly balanced non–tie choices (e.g., Gemma $39\!:\!36$, Llama $42\!:\!38$, Mistral $90\!:\!85$), indicating minimal label susceptibility in outcomes. Overall, revealing true labels sharply separates the schemes: Baseline $\gg$ SCoT in label anchoring, while PBP largely preserves Blind neutrality and tie behavior.

\subsection{Decisions with FLIP (Misleading) Labels in Summaries.}
Table~\ref{tab:FLIP-counts} presents raw counts when summaries carry \emph{FLIP} labels; the identification of candidates remains fixed (\(1\!=\!\)LLM, \(2\!=\!\)TradML). The \textbf{Baseline} again shows strong anchoring toward \([1,2]\) despite labels being incorrect (e.g., Gemma \(800{:}200\), Llama \(760{:}240\), Qwen \(750{:}250\), and \(1000{:}0\) for Mistral/Zephyr), with no abstention. \textbf{SCoT} introduces limited abstention (No Selection \(71\)–\(210\)) but still exhibits substantial label-directed preference (e.g., Gemma \(683{:}107\), Llama \(645{:}155\), Mistral \(898{:}31\), Qwen \(861{:}42\), Zephyr \(820{:}57\)), indicating susceptibility to misleading cues. In contrast, \textbf{PBP} largely preserves Blind behavior: high abstention (No Selection \(823\)–\(923\)) and non-tie choices that remain small and near-balanced (e.g., Gemma \(39{:}38\), Llama \(41{:}42\), Mistral \(91{:}86\), Qwen \(65{:}60\), Zephyr \(50{:}53\)). Overall, when exposed to incorrect labels, Baseline and SCoT continue to favor the label-indicated side, whereas PBP resists such anchoring by keeping most mass in \emph{Tie} and limiting outcome shifts.

\begin{table*}[!htbp]
\centering
\caption{FLIP-label decisions by scheme and model. Counts for \emph{No Selection} (Tie) and strict pairings \([1,2]\) (LLM \( \succ \) TradML) and \([2,1]\) (TradML \( \succ \) LLM). Despite labels being misleading, Baseline and SCoT still prefer \([1,2]\) strongly, while PBP maintains high abstention and near-balanced non-tie choices, indicating robustness to incorrect cues.}
\label{tab:FLIP-counts}
\begin{tabular}{|l|lll|lll|lll|}
\hline
\multirow{2}{*}{Model (Judge)} & \multicolumn{3}{l|}{Baseline}                                          & \multicolumn{3}{l|}{SCoT}                                              & \multicolumn{3}{l|}{PBP}                                               \\ \cline{2-10} 
                               & \multicolumn{1}{l|}{No Selection} & \multicolumn{1}{l|}{[1,2]} & [2,1] & \multicolumn{1}{l|}{No Selection} & \multicolumn{1}{l|}{[1,2]} & [2,1] & \multicolumn{1}{l|}{No Selection} & \multicolumn{1}{l|}{[1,2]} & [2,1] \\ \hline
Gemma-2-9B                     & \multicolumn{1}{l|}{0}            & \multicolumn{1}{l|}{800}   & 200   & \multicolumn{1}{l|}{210}          & \multicolumn{1}{l|}{683}   & 107   & \multicolumn{1}{l|}{923}          & \multicolumn{1}{l|}{39}    & 38    \\ \hline
Llama-3.1-8B                   & \multicolumn{1}{l|}{0}            & \multicolumn{1}{l|}{760}   & 240   & \multicolumn{1}{l|}{200}          & \multicolumn{1}{l|}{645}   & 155   & \multicolumn{1}{l|}{917}          & \multicolumn{1}{l|}{41}    & 42    \\ \hline
Mistral-7B                     & \multicolumn{1}{l|}{0}            & \multicolumn{1}{l|}{1000}  & 0     & \multicolumn{1}{l|}{71}           & \multicolumn{1}{l|}{898}   & 31    & \multicolumn{1}{l|}{823}          & \multicolumn{1}{l|}{91}    & 86    \\ \hline
Qwen2.5-7B                     & \multicolumn{1}{l|}{0}            & \multicolumn{1}{l|}{750}   & 250   & \multicolumn{1}{l|}{97}           & \multicolumn{1}{l|}{861}   & 42    & \multicolumn{1}{l|}{875}          & \multicolumn{1}{l|}{65}    & 60    \\ \hline
Zephyr-7B                      & \multicolumn{1}{l|}{0}            & \multicolumn{1}{l|}{1000}  & 0     & \multicolumn{1}{l|}{123}          & \multicolumn{1}{l|}{820}   & 57    & \multicolumn{1}{l|}{897}          & \multicolumn{1}{l|}{50}    & 53    \\ \hline
\end{tabular}
\end{table*}

\begin{table*}[!htbp]
\centering
\caption{Placebo–label decisions by scheme and model (\(1{=}\)LLM, \(2{=}\)TradML). Counts for \emph{No Selection} (Tie) and strict pairings $[1,2]$ and $[2,1]$. Placebo labels should be irrelevant; nonetheless, Baseline and SCoT exhibit sizable shifts toward one side (and a consistent $[2,1]$ preference for \emph{Qwen2.5-7B}), whereas PBP sustains high abstention and near-balanced non–ties, indicating robustness to irrelevant cues.}
\label{tab:placebo-counts}
\begin{tabular}{|l|lll|lll|lll|}
\hline
\multirow{2}{*}{Model (Judge)} & \multicolumn{3}{l|}{Baseline}                                          & \multicolumn{3}{l|}{SCoT}                                              & \multicolumn{3}{l|}{PBP}                                               \\ \cline{2-10} 
                               & \multicolumn{1}{l|}{No Selection} & \multicolumn{1}{l|}{[1,2]} & [2,1] & \multicolumn{1}{l|}{No Selection} & \multicolumn{1}{l|}{[1,2]} & [2,1] & \multicolumn{1}{l|}{No Selection} & \multicolumn{1}{l|}{[1,2]} & [2,1] \\ \hline
Gemma-2-9B                     & \multicolumn{1}{l|}{0}            & \multicolumn{1}{l|}{960}   & 40    & \multicolumn{1}{l|}{240}          & \multicolumn{1}{l|}{540}   & 220   & \multicolumn{1}{l|}{928}          & \multicolumn{1}{l|}{39}    & 33    \\ \hline
Llama-3.1-8B                   & \multicolumn{1}{l|}{0}            & \multicolumn{1}{l|}{680}   & 320   & \multicolumn{1}{l|}{213}          & \multicolumn{1}{l|}{620}   & 167   & \multicolumn{1}{l|}{917}          & \multicolumn{1}{l|}{44}    & 39    \\ \hline
Mistral-7B                     & \multicolumn{1}{l|}{0}            & \multicolumn{1}{l|}{1000}  & 0     & \multicolumn{1}{l|}{95}           & \multicolumn{1}{l|}{880}   & 25    & \multicolumn{1}{l|}{816}          & \multicolumn{1}{l|}{97}    & 92    \\ \hline
Qwen2.5-7B                     & \multicolumn{1}{l|}{0}            & \multicolumn{1}{l|}{230}   & 770   & \multicolumn{1}{l|}{103}          & \multicolumn{1}{l|}{127}   & 770   & \multicolumn{1}{l|}{879}          & \multicolumn{1}{l|}{62}    & 59    \\ \hline
Zephyr-7B                      & \multicolumn{1}{l|}{0}            & \multicolumn{1}{l|}{920}   & 80    & \multicolumn{1}{l|}{140}          & \multicolumn{1}{l|}{718}   & 142   & \multicolumn{1}{l|}{890}          & \multicolumn{1}{l|}{54}    & 56    \\ \hline
\end{tabular}
\end{table*}

\begin{table*}[!htbp]
\centering
\caption{Decomposition of outcome shifts from Blind to True when both candidates are LLM outputs (two independent samples), while labels still present $1{=}$LLM and $2{=}$TradML. $\mathrm{LDS}_T$ denotes movement toward the label–favored side $[1,2]$, $\mathrm{TS}_T$ movement into Tie, and $\mathrm{OLS}_T$ movement toward $[2,1]$. Large $\mathrm{LDS}_T$ with $\mathrm{TS}_T{\approx}0$ and $\mathrm{OLS}_T{\approx}0$ evidences pure label anchoring; PBP nearly eliminates it.}
\label{tab:llm-llm-true}
\begin{tabular}{|l|lll|lll|lll|}
\hline
\multirow{2}{*}{Model (Judge)} & \multicolumn{3}{c|}{Baseline}                                   & \multicolumn{3}{c|}{SCoT}                                       & \multicolumn{3}{c|}{PBP}                                        \\ \cline{2-10} 
                            & \multicolumn{1}{l|}{$LDS_T$} & \multicolumn{1}{l|}{$TS_T$}  & $OLS_T$ & \multicolumn{1}{l|}{$LDS_T$} & \multicolumn{1}{l|}{$TS_T$}  & $OLS_T$ & \multicolumn{1}{l|}{$LDS_T$} & \multicolumn{1}{l|}{$TS_T$}  & $OLS_T$ \\ \hline
Gemma-2-9B                     & \multicolumn{1}{l|}{0.509} & \multicolumn{1}{l|}{0.000} & 0.000 & \multicolumn{1}{l|}{0.583} & \multicolumn{1}{l|}{0.000} & 0.041 & \multicolumn{1}{l|}{0.005} & \multicolumn{1}{l|}{0.000} & 0.007 \\ \hline
Llama-3.1-8B                   & \multicolumn{1}{l|}{0.390} & \multicolumn{1}{l|}{0.000} & 0.000 & \multicolumn{1}{l|}{0.550} & \multicolumn{1}{l|}{0.000} & 0.056 & \multicolumn{1}{l|}{0.005} & \multicolumn{1}{l|}{0.000} & 0.005 \\ \hline
Mistral-7B                     & \multicolumn{1}{l|}{0.450} & \multicolumn{1}{l|}{0.000} & 0.000 & \multicolumn{1}{l|}{0.714} & \multicolumn{1}{l|}{0.000} & 0.000 & \multicolumn{1}{l|}{0.000} & \multicolumn{1}{l|}{0.004} & 0.000 \\ \hline
Qwen2.5-7B                     & \multicolumn{1}{l|}{0.290} & \multicolumn{1}{l|}{0.000} & 0.000 & \multicolumn{1}{l|}{0.699} & \multicolumn{1}{l|}{0.000} & 0.000 & \multicolumn{1}{l|}{0.010} & \multicolumn{1}{l|}{0.000} & 0.003 \\ \hline
Zephyr-7B                      & \multicolumn{1}{l|}{0.490} & \multicolumn{1}{l|}{0.000} & 0.000 & \multicolumn{1}{l|}{0.583} & \multicolumn{1}{l|}{0.000} & 0.041 & \multicolumn{1}{l|}{0.001} & \multicolumn{1}{l|}{0.000} & 0.001 \\ \hline
\end{tabular}
\end{table*}

\begin{table*}[!htbp]
\centering
\caption{Decomposition of outcome shifts from Blind under \emph{FLIP} labels when both candidates are LLM outputs (two independent draws). $\mathrm{LDS}_F$ = movement toward $[1,2]$ (label–favored); $\mathrm{TS}_F$ = into Tie; $\mathrm{OLS}_F$ = toward $[2,1]$. PBP $\approx 0$ on all components; Baseline moves slightly opposite the cue; SCoT shows mixed behavior with large label–directed shifts for some models (e.g., Qwen).}
\label{tab:lds-ts-ols-FLIP}
\begin{tabular}{|l|lll|lll|lll|}
\hline
\multirow{2}{*}{Model (Judge)} & \multicolumn{3}{c|}{Baseline}                                   & \multicolumn{3}{c|}{SCoT}                                       & \multicolumn{3}{c|}{PBP}                                        \\ \cline{2-10} 
                               & \multicolumn{1}{l|}{$LDS_F$} & \multicolumn{1}{l|}{$TS_F$}  & $OLS_F$ & \multicolumn{1}{l|}{$LDS_F$} & \multicolumn{1}{l|}{$TS_F$}  & $OLS_F$ & \multicolumn{1}{l|}{$LDS_F$} & \multicolumn{1}{l|}{$TS_F$}  & $OLS_F$ \\ \hline
Gemma-2-9B                     & \multicolumn{1}{l|}{0.000} & \multicolumn{1}{l|}{0.000} & 0.240 & \multicolumn{1}{l|}{0.044} & \multicolumn{1}{l|}{0.000} & 0.142 & \multicolumn{1}{l|}{0.009} & \multicolumn{1}{l|}{0.000} & 0.005 \\ \hline
Llama-3.1-8B                   & \multicolumn{1}{l|}{0.000} & \multicolumn{1}{l|}{0.000} & 0.230 & \multicolumn{1}{l|}{0.081} & \multicolumn{1}{l|}{0.000} & 0.248 & \multicolumn{1}{l|}{0.003} & \multicolumn{1}{l|}{0.000} & 0.004 \\ \hline
Mistral-7B                     & \multicolumn{1}{l|}{0.000} & \multicolumn{1}{l|}{0.000} & 0.250 & \multicolumn{1}{l|}{0.000} & \multicolumn{1}{l|}{0.000} & 0.028 & \multicolumn{1}{l|}{0.000} & \multicolumn{1}{l|}{0.006} & 0.000 \\ \hline
Qwen2.5-7B                     & \multicolumn{1}{l|}{0.000} & \multicolumn{1}{l|}{0.000} & 0.250 & \multicolumn{1}{l|}{0.666} & \multicolumn{1}{l|}{0.000} & 0.000 & \multicolumn{1}{l|}{0.003} & \multicolumn{1}{l|}{0.000} & 0.007 \\ \hline
Zephyr-7B                      & \multicolumn{1}{l|}{0.000} & \multicolumn{1}{l|}{0.000} & 0.210 & \multicolumn{1}{l|}{0.019} & \multicolumn{1}{l|}{0.000} & 0.225 & \multicolumn{1}{l|}{0.000} & \multicolumn{1}{l|}{0.003} & 0.000 \\ \hline
\end{tabular}
\end{table*}
\begin{table*}[!htbp]
\centering
\caption{Decomposition of outcome shifts from Blind under \emph{placebo} labels with both candidates drawn from the same LLM. $\mathrm{LDS}_P$ = movement toward the cue-favored outcome $[1,2]$; $\mathrm{TS}_P$ = into \emph{Tie}; $\mathrm{OLS}_P$ = toward $[2,1]$. PBP $\approx 0$ across components (best); Baseline shows moderate placebo anchoring; SCoT is most susceptible.}
\label{tab:lds-ts-ols-placebo}
\begin{tabular}{|l|lll|lll|lll|}
\hline
\multirow{2}{*}{Model (Judge)} & \multicolumn{3}{c|}{Baseline}                                   & \multicolumn{3}{c|}{SCoT}                                       & \multicolumn{3}{c|}{PBP}                                        \\ \cline{2-10} 
                               & \multicolumn{1}{l|}{$LDS_P$} & \multicolumn{1}{l|}{$TS_P$}  & $OLS_P$ & \multicolumn{1}{l|}{$LDS_P$} & \multicolumn{1}{l|}{$TS_P$}  & $OLS_P$ & \multicolumn{1}{l|}{$LDS_P$} & \multicolumn{1}{l|}{$TS_P$}  & $OLS_T$ \\ \hline
Gemma-2-9B                     & \multicolumn{1}{l|}{0.420} & \multicolumn{1}{l|}{0.000} & 0.000 & \multicolumn{1}{l|}{0.586} & \multicolumn{1}{l|}{0.000} & 0.279 & \multicolumn{1}{l|}{0.005} & \multicolumn{1}{l|}{0.000} & 0.005 \\ \hline
Llama-3.1-8B                   & \multicolumn{1}{l|}{0.180} & \multicolumn{1}{l|}{0.000} & 0.084 & \multicolumn{1}{l|}{0.560} & \multicolumn{1}{l|}{0.000} & 0.302 & \multicolumn{1}{l|}{0.005} & \multicolumn{1}{l|}{0.000} & 0.004 \\ \hline
Mistral-7B                     & \multicolumn{1}{l|}{0.490} & \multicolumn{1}{l|}{0.000} & 0.000 & \multicolumn{1}{l|}{0.688} & \multicolumn{1}{l|}{0.000} & 0.236 & \multicolumn{1}{l|}{0.000} & \multicolumn{1}{l|}{0.005} & 0.000 \\ \hline
Qwen2.5-7B                     & \multicolumn{1}{l|}{0.000} & \multicolumn{1}{l|}{0.270} & 0.000 & \multicolumn{1}{l|}{0.000} & \multicolumn{1}{l|}{0.000} & 0.576 & \multicolumn{1}{l|}{0.010} & \multicolumn{1}{l|}{0.000} & 0.006 \\ \hline
Zephyr-7B                      & \multicolumn{1}{l|}{0.420} & \multicolumn{1}{l|}{0.000} & 0.000 & \multicolumn{1}{l|}{0.587} & \multicolumn{1}{l|}{0.000} & 0.326 & \multicolumn{1}{l|}{0.000} & \multicolumn{1}{l|}{0.004} & 0.000 \\ \hline
\end{tabular}
\end{table*}
\subsection{Decisions with Placebo Labels (Placebo labels, $1{=}$LLM, $2{=}$TradML).}
Table~\ref{tab:placebo-counts} reports raw outcomes when summaries carry \emph{placebo} labels that should be ignored. A robust judge ought to preserve its Blind behavior, i.e., maintain high abstention and near-balanced non–tie choices. The \textbf{Baseline} fails this desideratum: it never abstains and remains strongly polarized toward $[1,2]$ for most models (e.g., Gemma $960{:}40$, Mistral $1000{:}0$, Zephyr $920{:}80$), with a notable reversal for \emph{Qwen2.5-7B} ($230{:}770$ toward $[2,1]$), indicating substantial sensitivity even to Placebo labels. \textbf{SCoT} introduces some abstention (No Selection $95$–$240$) yet still displays large placebo-driven preferences (e.g., Gemma $540{:}220$, Llama $620{:}167$, Mistral $880{:}25$), and again markedly favors $[2,1]$ for \emph{Qwen2.5-7B} ($127{:}770$), consistent with strong cue susceptibility. In contrast, \textbf{PBP} largely preserves Blind neutrality: abstention remains high (No Selection $816$–$928$) and the residual non–tie decisions are both small and nearly balanced (e.g., Gemma $39{:}33$, Llama $44{:}39$, Mistral $97{:}92$, Qwen $62{:}59$, Zephyr $54{:}56$). These counts align with the tie–aware, directional analysis (\(\mathrm{LAO}_{\text{Placebo}}\) low for PBP, high for Baseline/SCoT): placebo labels spur outcome shifts for Baseline and SCoT but are largely defused by PBP’s evidence–first workflow.


\subsection{Label anchoring when both candidates are LLM (two independent draws).}
To isolate \emph{pure} label effects, we compare two summaries sampled independently from the same LLM (candidate $1$ and $2$ are both LLM outputs) while keeping the display labels fixed as $1{=}$LLM and $2{=}$TradML. We decompose the change from Blind to True into \emph{label–directed shift} ($\mathrm{LDS}_T$), \emph{movement into Tie} ($\mathrm{TS}_T$), and \emph{movement toward the opposite side} ($\mathrm{OLS}_T$). As Table~\ref{tab:llm-llm-true} shows, the \textbf{Baseline} channels a large fraction of probability mass toward the label–favored decision $[1,2]$ (e.g., Gemma $0.509$, Llama $0.390$, Mistral $0.450$, Qwen $0.290$, Zephyr $0.490$) with essentially no countervailing movement ($\mathrm{TS}_T{=}0$, $\mathrm{OLS}_T{=}0$), indicating strong label anchoring even though the candidates are content–equivalent in provenance. \textbf{SCoT} amplifies this effect: $\mathrm{LDS}_T$ is larger than Baseline for every model (up to $0.714$ for Mistral), while $\mathrm{OLS}_T$ is only marginally nonzero ($0.041$–$0.056$), yielding a net pull toward the label–favored outcome. In contrast, \textbf{PBP} substantially suppresses label influence: $\mathrm{LDS}_T$ is near zero (Gemma $0.005$, Llama $0.005$, Mistral $0.000$, Qwen $0.010$, Zephyr $0.001$), with small compensating moves into Tie or the opposite side (e.g., Mistral $\mathrm{TS}_T{=}0.004$), effectively preserving Blind behavior. Because the two candidates are generated by the \emph{same} model, these shifts cannot be attributed to content quality; rather, they quantify \emph{rationalization bias} driven solely by the label framing, with PBP offering the strongest mitigation, Baseline moderate susceptibility, and SCoT the highest susceptibility.


\begin{table*}[!htbp]
\centering
\caption{Head-to-head decisions for candidate $1$ (LLM) versus candidate $2$ (TradML) across four extractive baselines. Entries are counts of $[1,2]$ (LLM $\succ$ TradML) and $[2,1]$ (TradML $\succ$ LLM). The LLM dominates universally, with only two minor exceptions (Llama vs.\ SumBasic: $995{:}5$; Qwen vs.\ LexRank: $970{:}30$).}
\label{tml}
\begin{tabular}{|l|ll|ll|ll|ll|}
\hline
\multirow{2}{*}{Model (Judge)} & \multicolumn{2}{l|}{LexRank}       & \multicolumn{2}{l|}{TextRank}      & \multicolumn{2}{l|}{KL-Sum}        & \multicolumn{2}{l|}{SumBasic}      \\ \cline{2-9} 
                               & \multicolumn{1}{l|}{[1,2]} & [2,1] & \multicolumn{1}{l|}{[1,2]} & [2,1] & \multicolumn{1}{l|}{[1,2]} & [2,1] & \multicolumn{1}{l|}{[1,2]} & [2,1] \\ \hline
Gemma-2-9B                     & \multicolumn{1}{l|}{1000}  & 0     & \multicolumn{1}{l|}{1000}  & 0     & \multicolumn{1}{l|}{1000}  & 0     & \multicolumn{1}{l|}{1000}  & 0     \\ \hline
Llama-3.1-8B                   & \multicolumn{1}{l|}{1000}  & 0     & \multicolumn{1}{l|}{1000}  & 0     & \multicolumn{1}{l|}{1000}  & 0     & \multicolumn{1}{l|}{995}   & 5     \\ \hline
Mistral-7B                     & \multicolumn{1}{l|}{1000}  & 0     & \multicolumn{1}{l|}{1000}  & 0     & \multicolumn{1}{l|}{1000}  & 0     & \multicolumn{1}{l|}{1000}  & 0     \\ \hline
Qwen2.5-7B                     & \multicolumn{1}{l|}{970}   & 30    & \multicolumn{1}{l|}{1000}  & 0     & \multicolumn{1}{l|}{1000}  & 0     & \multicolumn{1}{l|}{1000}  & 0     \\ \hline
Zephyr-7B                      & \multicolumn{1}{l|}{1000}  & 0     & \multicolumn{1}{l|}{1000}  & 0     & \multicolumn{1}{l|}{1000}  & 0     & \multicolumn{1}{l|}{1000}  & 0     \\ \hline
\end{tabular}
\end{table*}

\begin{table*}[!htbp]
\centering
\caption{Identical-summary test: raw counts when both candidates contain the \emph{same} text but are labeled $1$ (LLM) and $2$ (TradML). The very high abstention and near-balanced rare picks indicate minimal bias when content is controlled.}
\label{tab:identical}
\begin{tabular}{|l|l|l|l|}
\hline
Model (Judge) & No Selection & [1, 2] & [2,1] \\ \hline
Gemma-2-9B    & 987          & 6      & 7     \\ \hline
Llama-3.1-8B  & 985          & 9      & 6     \\ \hline
Mistral-7B    & 976          & 13     & 11    \\ \hline
Qwen2.5-7B    & 982          & 10     & 8     \\ \hline
Zephyr-7B     & 981          & 9      & 10    \\ \hline
\end{tabular}
\end{table*}
\subsection{FLIP-Label Outcome Shift with Same-Model Candidates (LDS/TS/OLS).}
Table~\ref{tab:lds-ts-ols-FLIP} decomposes the change from Blind when \emph{FLIP} (misleading) labels are displayed, for the case in which both candidates are independently sampled from the same LLM (labels fixed as $1{=}$LLM, $2{=}$TradML). We report $\mathrm{LDS}_F$ (positive movement toward the label–favored side $[1,2]$), $\mathrm{TS}_F$ (into Tie), and $\mathrm{OLS}_F$ (toward the opposite side $[2,1]$). The \textbf{Baseline} shows \emph{no} label–ward movement ($\mathrm{LDS}_F{=}0$ across models) and a moderate shift \emph{against} the misleading cue ($\mathrm{OLS}_F\!\approx\!0.21$–$0.25$), with no mass going to Tie ($\mathrm{TS}_F{=}0$). \textbf{SCoT} is mixed but notably vulnerable: while some models still move opposite the cue (e.g., Gemma $\mathrm{OLS}_F{=}0.142$, Llama $0.248$, Zephyr $0.225$), others exhibit substantial \emph{label–directed} movement despite the labels being incorrect (e.g., Qwen $\mathrm{LDS}_F{=}0.666$; Llama $\mathrm{LDS}_F{=}0.081$). In contrast, \textbf{PBP} keeps all components near zero ($\mathrm{LDS}_F\!\le\!0.009$, $\mathrm{TS}_F\!\le\!0.006$, $\mathrm{OLS}_F\!\le\!0.007$), effectively preserving Blind behavior under misleading cues. Overall, these results confirm that PBP minimizes framing-only rationalization under FLIP labels, Baseline exhibits modest corrective movement away from the cue, and SCoT remains the most susceptible to misleading label pressure.


\subsection{Placebo-Label Outcome Shift with Same-Model Candidates (LDS/TS/OLS).}
Table~\ref{tab:lds-ts-ols-placebo} decomposes the change from Blind when \emph{placebo} (irrelevant) labels are displayed, holding provenance fixed by sampling both candidates from the same LLM ($1{=}$LLM, $2{=}$TradML). We report $\mathrm{LDS}_P$ (positive movement toward the cue-favored side $[1,2]$), $\mathrm{TS}_P$ (movement into \emph{Tie}), and $\mathrm{OLS}_P$ (movement toward the opposite side $[2,1]$). Ideally, placebo cues should be ignored, yielding $\mathrm{LDS}_P\!\approx\!0$ and at most small $\mathrm{TS}_P$ (benign) or $\mathrm{OLS}_P$ (counter-cue). The \textbf{Baseline} shows moderate placebo anchoring with sizeable $\mathrm{LDS}_P$ for most judges (Gemma $0.420$, Mistral $0.490$, Zephyr $0.420$; Llama $0.180$ plus some opposite shift $\mathrm{OLS}_P{=}0.084$), while \emph{Qwen2.5-7B} pushes mass into \emph{Tie} ($\mathrm{TS}_P{=}0.270$), reflecting partial cue immunity. \textbf{SCoT} is markedly susceptible: $\mathrm{LDS}_P$ is large across models (Gemma $0.586$, Llama $0.560$, Mistral $0.688$, Zephyr $0.587$), with nontrivial opposite shifts for some (e.g., Gemma $\mathrm{OLS}_P{=}0.279$, Llama $0.302$, Zephyr $0.326$), and a striking counter-cue for \emph{Qwen2.5-7B} ($\mathrm{OLS}_P{=}0.576$) but \emph{no} move into \emph{Tie}. In contrast, \textbf{PBP} remains near Blind: all components are tiny ($\mathrm{LDS}_P\le 0.010$, $\mathrm{TS}_P\le 0.005$, $\mathrm{OLS}_P\le 0.006$), indicating that Placebo labels are largely ignored. Overall, placebo labels elicit outcome shifts for Baseline and especially SCoT, whereas PBP effectively neutralizes them.


\subsection{Sanity check with identical summaries (labels only).}
To isolate pure label effects, we present \emph{identical} summaries while assigning them different labels ($1{=}$LLM, $2{=}$TradML). As shown in Table~\ref{tab:identical}, judges overwhelmingly abstain (\emph{No Selection} $=\!$ 976–987/1000), yielding very high equality detection ($\mathrm{EDR}\!\approx\!0.976$–$0.987$). The residual non–tie choices are rare and nearly balanced (e.g., Gemma 6 vs.\ 7; Llama 9 vs.\ 6; Zephyr 9 vs.\ 10), implying a near–zero neutrality deviation and confirming that, when content is strictly controlled, decisions are not driven by the label identity. This sanity check supports our interpretation of label–anchoring results: outcome shifts observed under \textsc{True}/\textsc{Flip}/\textsc{Placebo} arise from cue susceptibility rather than an inherent preference for the LLM (1) or TradML (2) label.


\subsection{LLM vs.\ Classical Extractive Baselines.}
We compare candidate $1$ (LLM) against candidate $2$ (TradML) across four extractive baselines, \emph{LexRank}, \emph{TextRank}, \emph{KL-Sum}, and \emph{SumBasic}. Across judges and datasets, the preference is overwhelmingly in favor of the LLM summaries. For \textit{Gemma-2-9B}, \textit{Mistral-7B}, and \textit{Zephyr-7B}, the LLM is selected in \emph{all} evaluations for all four baselines ($[1,2]=1000$, $[2,1]=0$). \textit{Llama-3.1-8B} shows the same unanimity for LexRank, TextRank, and KL-Sum ($1000{:}0$ each), with a single minor deviation against SumBasic ($995{:}5$). \textit{Qwen2.5-7B} is likewise unanimous for TextRank, KL-Sum, and SumBasic ($1000{:}0$), and remains decisively in favor of the LLM against LexRank ($970{:}30$). These results indicate a consistent and substantive quality advantage of LLM-generated summaries over classical extractive methods, rather than an effect of label susceptibility: when content differs, judges nearly always prefer the LLM output.

\end{quote}



\end{document}